\documentclass[journal]{IEEEtran}
\usepackage{amsfonts,amssymb, amsthm,amsmath}
\usepackage[english]{babel}
\usepackage{subeqnarray}
\usepackage{cases}
\usepackage{tikz}
\usepackage{graphicx}
\usepackage{float}
\usepackage[labelfont=bf,labelsep=space]{caption}
\usepackage{color}
\usepackage{tikz}
\usepackage{arydshln}
\usepackage{stfloats}
\usepackage{cite}
\usepackage{amsfonts}
\usepackage{enumerate}
\usepackage{bm}
\usepackage[ruled]{algorithm2e}
\usepackage{url}
\usepackage{hyperref}
\usepackage[utf8]{inputenc}
\usepackage{xcolor}
\hypersetup{
    colorlinks=true,
    linkcolor=black,
    filecolor=magenta,    
    citecolor=black,
    urlcolor=blue,
    }
\ifCLASSINFOpdf
\else

\fi
\newtheorem{theorem}{Theorem}

\newtheorem{lemma}{Lemma}

\newtheorem{remark}{Remark}
\newtheorem{assumption}{Assumption}

\newcommand{\V}{\mathbf V}
\newcommand{\E}{\mathbf E}

\newcommand{\trace}{\mathbf{trace}}
\newcommand{\Var}{\mathbf{Var}}

\newcommand{\step}{\mathrm{step}}

\newcommand{\si}{\mathrm{si}}
\newcommand{\argmax}{\mathrm{argmax}}

\def\bs{\boldsymbol s}
\def\bu{\boldsymbol u}
\def\bv{\boldsymbol v}
\def\bx{\boldsymbol x}
\def\bp{\boldsymbol p}
\def\bs{\boldsymbol s}
\def\be{\boldsymbol e}
\def\bp{\boldsymbol p}
\def\bo{\boldsymbol o}

\def\by{\boldsymbol y}
\def\b\alpha{\boldsymbol \alpha}

\def\diag{\mathsf{diag}}

\begin{document}

\title{Distance-based Multiple Non-cooperative Ground Target Encirclement for Complex Environments
}

\author{
Fen Liu, Shenghai Yuan, Kun Cao, Wei Meng, Lihua Xie, \emph{Fellow, IEEE}

 \thanks{This research is supported by the National Research Foundation, Singapore under its Medium Sized Center for Advanced Robotics Technology Innovation, the National Natural Science Foundation of China (U21A20476, 62121004), and Shanghai Municipal Science and Technology Major Project (No. 2021SHZDZX0100). (Corresponding author:
Shanghai Yuan.)}

\thanks{F. Liu, S. Yuan, and L. Xie are with the School of Electrical and Electronic Engineering, Nanyang Technological University, Singapore 639798, Singapore (e-mail:fen.liu@ntu.edu.sg, shyuan@ntu.edu.sg, elhxie@ntu.edu.sg).

K. Cao is with the Department of Control Science and Engineering, College of Electronics and Information Engineering, and Shanghai Research Institute for Intelligent Autonomous Systems, Tongji University, Shanghai 200000, China (e-mail:caokun@tongji.edu.cn).

 W. Meng is with Guangdong Provincial Key Laboratory of Intelligent Decision and Cooperative Control, School of Automation, Guangdong University of Technology, Guangzhou 510006, China (e-mail: meng0025@ntu.edu.sg).}
}

\markboth{IEEE Transactions on Control Systems Technology }%
{Shell \MakeLowercase{\textit{et al.}}: Bare Demo of IEEEtran.cls for Journals}

\maketitle
\begin{abstract}
This paper proposes a comprehensive strategy for complex multi-target-multi-drone encirclement in an obstacle-rich and GPS-denied environment, motivated by practical scenarios such as pursuing vehicles or humans in urban canyons. The drones have omnidirectional range sensors that can robustly detect ground targets and obtain noisy relative distances. After each drone task is assigned, a novel distance-based target state estimator (DTSE) is proposed by estimating the measurement output noise variance and utilizing the Kalman filter.
By integrating anti-synchronization techniques and pseudo-force functions, an acceleration controller enables two tasking drones to cooperatively encircle a target from opposing positions while navigating obstacles.
The algorithm’s effectiveness for the discrete-time double-integrator system is established theoretically, 
particularly regarding observability.
Moreover, the versatility of the algorithm is showcased in aerial-to-ground scenarios, supported by compelling simulation results. Experimental validation demonstrates the effectiveness of the proposed approach.
\end{abstract}

\begin{IEEEkeywords} Multi-target-Multi-drone, Encirclement, Obstacle-rich, Distance-based target state estimator (DTSE).
\end{IEEEkeywords}

\IEEEpeerreviewmaketitle
\section{Introduction}
In recent years, the application of encirclement control has witnessed remarkable growth across diverse fields \cite{dong2019flight,hafez2015solving,sun2018circular}, including collaborative maritime target tracking \cite{hungrange,bayatrange,hungcooperative}, criminal chasing in law enforcement operations \cite{jin2019dynamic,deng2022gaitfi}, collaborative aerial inspections \cite{lyu2022structure, krizmancic2020cooperative}, exploration and rescue operations for the civil defense operations\cite{lee2021upper}.
Under encirclement control, drones can effectively track, monitor, and even capture targets. However, in certain situations, encirclement control encounters additional complexities, including denied GPS, uncooperative targets, and the presence of both mobile and stationary obstacles. 


In most existing encirclement studies, the consideration revolves around the ability to acquire the absolute \cite{yang2023av} or relative position of the target \cite{kou2021cooperative,aranda2014three,peng2019cooperative}. Map-based or GPS-based methods typically require information exchange between the target and the tracking drones, e.g., feature points and landmarks \cite{choudhary2017distributed}.
Various localization algorithms  \cite{deghat2014localization,liu2023non,nguyen2019single,li2022three,chen2022triangular,wang2023bearing} based on angle, bearing, and distance have been developed for non-cooperative targets that do not provide active feedback.
With only one single measurement, neither static nor moving targets can be accurately positioned.
Therefore, the existing
localization algorithms typically require sufficient measurements for positioning, e.g., multiple relative distances or bearings to the target through the ego-movements of a single drone \cite{li2022three,nguyen2019persistently,jiang2016simultaneous}, diverse types of measurements by multiple sensors on a single drone \cite{fang2023distributed}, multiple relative bearings or angles simultaneously measured by multiple drones  \cite{chen2022triangular,zhao2016localizability}, etc. The method of compensating for target localization through ego-movements typically does not work well for fast-moving or randomly moving targets \cite{shames2011circumnavigation}.  
Increasing the input dimensionality often requires more drones or sensors, increasing system costs and complexity. 
Algorithms that rely solely on minimal simultaneous measurements are of particular interest for further investigation.

While LiDAR provides both range and bearing measurements, it suffers from poor spatial resolution for general 3D object sensing. Furthermore, its mounting position above drones for obstacle detection leads to significant occlusion issues.
Gimbalized cameras can also be used for bearing and range measurements. However, when used for bearing measurements, they must be fused with a compass or IMU to provide a global heading reference \cite{bjphirneestimating}. This fusion can lead to inaccuracies over time due to IMU drift or electromagnetic interference, resulting in large uncertainties in angle measurements.
In contrast, using cameras for distance measurements simplifies the process by only requiring the detection of the relative size of the target without needing an angle reference \cite{vajgldist}. This method can be further enhanced by coupling it with cellphone signal strength or other wireless strength indicators to improve range measurements \cite{cairobust}. 
Range measurements are generally more practical and robust, as they are less prone to cumulative errors and environmental interference. 

Furthermore, when implementing encirclement control algorithms within complex obstacle-rich environments, integrating obstacle avoidance algorithms is very important. This integration ensures that drone systems can navigate around obstacles while achieving the designated encirclement objectives \cite{zhang2022multi}. 
To the best of our knowledge, most existing encirclement control algorithms have not considered obstacle avoidance \cite{kou2021cooperative, sun2018circular, peng2020event}. To avoid obstacles, a variety of existing algorithms in path planning-based algorithms can be considered, including potential field methods \cite{pan2021improved, yan2012multilevel}, graph-based search algorithms \cite{xiong2019path}, Monte-Carlo methods \cite{zucker2013chomp}, and ray-tracing methods \cite{zhang2020path}.
The integration of these methods with the encirclement control algorithm increases the complexity of overall algorithm implementation and optimization. \\
\indent To address these challenges, we focus on developing strategies for encircling swarms of randomly moving non-cooperative targets in obstacle-rich environments. 
The problem consists of several smaller subproblems, including task assignment for the drone team, estimation of non-cooperative target states, obstacle avoidance, and target encirclement. For the task assignment, we utilize the consensus-based auction algorithm (CBAA) \cite{choi2009consensus}. Then, a novel approach is proposed for estimating the state of moving ground targets with two synchronized drones, incorporating the Kalman filter and noise variance estimation techniques. Compared to existing works \cite{nguyen2019single,shames2011circumnavigation}, our approach effectively converts noisy distance measurements into position estimates, which are closer to real-world scenarios. Furthermore,  
we propose a novel acceleration controller by leveraging anti-synchronization (AS) \cite{liu2023multiple, liu2023moving} with an effective pseudo-force field. Compared to existing encirclement algorithms \cite{hafez2015solving,zhang2021decentralized}, this controller can efficiently enable drones to encircle a target in opposite directions while simultaneously avoiding obstacles. 
Meanwhile, the utilization of AS techniques is beneficial for the drone's observation of targets. Finally, in comparison to other localization and encirclement works \cite{li2022three, dong2019flight,sun2018circular}, we provide a rigorous analysis of observability and encirclement stability for discrete-time double-integrator systems in the presence of obstacles. A physical experiment involving UAVs and UGVs is designed to demonstrate the practicality of our proposed encirclement algorithms and highlight our contribution and impact in advancing real-world, cooperative drone applications. Detailed URL link for video demonstration can be found in \url{https://youtu.be/HAtOTANfCdY}.

This paper is organized as follows: Sec. II covers preliminaries and problem formulation. Sec. III introduces state estimation and control law. Sec. IV presents the convergence analysis. Sec. V provides numerical simulations and physical implementations, and Sec. VI concludes the article.\\
\indent Notations: The mean and the variance of a random variable are denoted as $\E{}$ and $\Var{}$, respectively. $\lambda \{A\}$ denotes an eigenvalue of matrix $A$, and the largest and smallest eigenvalues are denoted as $\lambda_{\max}\{A\}$ and $\lambda_{\min}\{A\}$, respectively. The transpose of matrix $A$ is represented by $A^\top $, while $A^{-1}$ signifies the inverse of matrix $A$. $I_{3\times 3}$ and $0_{3\times 3}$ are the $n$-dimensional identity and zero matrices, respectively. The Euclidean norm is denoted as $||\cdot||$, and $\max\{\}$ and $\min\{\}$ are the functions that select the maximum value and the minimum value, respectively. $\angle$ represents the included angle. Lastly, $[d_{ij}]_{n\times m}$ denotes the $n\times m$ matrix of elements $d_{ij}$.

\section{Preliminaries and Problem Formulation}
\subsection{System Model} \label{System_Model}
We consider a scenario involving $M$ ground moving targets, $N$ tasking drones and $O$ static obstacles, defined as the sets $\Phi_M\triangleq\{1,2,\ldots,M\}$, $\Phi_N\triangleq\{1,2,\ldots,N\}$, and $\Phi_O\triangleq\{1,2,\ldots,O\}$, respectively. 

Denote $\bx_i^{(k)} \in \mathbb{R}^3,$ and $\bv_i ^{(k)}\in \mathbb{R}^3, i \in \Phi_N$ as the position and velocity of the drone $i$ in the sampling instant $k$, respectively. The following discrete double-integrator model can describe the model of the drone $i$,
\begin{subequations}\label{eq1}
\begin{align}
\bx_i^{(k+1)}=&\bx_i^{(k)}+t\bv_i^{(k)}+\frac{1}{2}t^2\bu_i^{(k)},\label{eq1-1}\\
\bv_i^{(k+1)}=&\bv_i^{(k)}+t\bu_i^{(k)},\label{eq1-2}
\end{align}
\end{subequations}
where $t$ denotes the sampling period and $t>0$, and $\bu_i\in \mathbb{R}^3$ is the controlled acceleration. For the sake of simplification, denote $\bx_{i}^{(-)}=\bx_{i}(k-1)$, $\bx_{i}=\bx_{i}^{(k)}$, and $\bx_{i}^{(+)}=\bx_{i}^{(k+1)}$.

By augmenting the state $\bm{\xi}_i=[\bx_i,\bv_i]^\top  \in \mathbb{R}^{6}$, the model of drone $i$ can be rewritten as
\begin{equation}\label{eq2}
\begin{split}
\bm{\xi}_i^{(+)}=A_3\bm{\xi}_i+B_3\bu_i,
\end{split}
\end{equation}
where
\begin{equation*}
\begin{split}
A_3=\left[
    \begin{array}{cc}
      I_{3\times 3} &tI_{3\times 3} \\
      0_{3\times 3} & I_{3\times 3} \\
    \end{array}
  \right],~B_3=\left[
    \begin{array}{c}
      \frac{1}{2}t^2I_{3\times 3}\\
      tI_{3\times 3} \\
    \end{array}
  \right].
\end{split}
\end{equation*}
We denote $\lambda_{\min}\{A_3A_3^\top\}=\underline{a}$, $\lambda_{\max}\{A_3A_3^\top\}=\bar{a}$, $\lambda_{\min}\{B_3B_3^\top\}=\underline{b}$ and $\lambda_{\max}\{B_3B_3^\top\}=\bar{b}$. Through simple derivation, we have $\underline{a}= \frac{2+t^2-t\sqrt{4+t^2}}{2}$, $\bar{a}=\frac{2+t^2+ t\sqrt{4+t^2}}{2}$, $\underline{b}=0$ and $\bar{b}=\frac{1}{4}t^4+t^2$.

For non-cooperative targets, we assume that the real-time state information is unavailable by communication, and the movements are dispersed.
Denote $\bm{\eta}_j=[\bs_j,\bm{\nu}_j] \in \mathbb{R}^{4},j \in \Phi_M$ as the state of the target $j$, where $\bs_j$ and $\bm{\nu}_j$ are the position and the velocity of target $j$, respectively. Assume the state model of target $j$ as
\begin{equation}\label{eq3}
\begin{split}
\bm{\eta}_j^{(+)}=A_2\bm{\eta}_j+B_2\bm{\omega}_j, 
\end{split}
\end{equation}
where $\bm{\omega}_j \in \mathbb{R}^2$ is the stochastic acceleration and satisfies zero-mean Gaussian distribution, i.e., $\E\{\bm{\omega}_j\}=0$ and $\Var\{\bm{\omega}_j\}=Q_j$, where $Q_j \in \mathbb{R}^2$ is the symmetric positive definite matrix. 
\begin{remark}\label{input noise variance}
Obtaining the real-time states of a non-cooperative target is challenging.
However, for common types of targets, some specific characteristics of acceleration can be anticipated in advance, such as the maximum acceleration of a certain type of car. Hence, in this study, we assume that the target motion acceleration variance is known.
\end{remark}
Most obstacles are static in the real world, especially when employing aerial drones to encircle ground targets. Therefore, we only consider the static obstacles in this work. The real-time position of static obstacle $\iota$ is defined as $\bo_\iota \in \mathbb{R}^{3}, \iota  \in \Phi_O$, which can be detected by the onboard sensor of drones. Additionally, the drones follow these assumptions:

\begin{assumption} \label{Number of drones and targets} (Number of drones and targets)
The number of tasking drones is set to be twice the number of targets, i.e., $N = 2M$. According to the task assignment algorithm, all drones are divided into $M$ groups.
\end{assumption}

\begin{assumption} \label{Radius of communication and measurement} (Radius of communication and measurement)
The communication radius and measurement radius of each drone are $r_1$ and $r_2$, respectively. Supposing communication exists between drones capable of measuring the same target, we have $r_1 \geq 2r_2$.
\end{assumption}


\begin{assumption} \label{Initial coordinate system matching} (Initial coordinate system matching) 
In the initial phase, each drone is presumed to be able to observe nearby landmarks and share their relative positions with others in one group, enabling them to have the same coordinate system.
\end{assumption}

Let $\bp_{i,g}$ be the relative position between drone $i$ and drone $g$ in same coordinate
system, i.e., $\bp_{i,g}=\bx_{i}-\bx_{g}$. Based on the Assumption \ref{Radius of communication and measurement}, we have $\bp_{i,g}\bp_{i,g}^{\top}\leq \check{c}I_{3\times 3}$ with positive real constant $\check{c}$ if drones $i$ and $g$ are in one group.

\begin{figure}
\centering
  \includegraphics[width=7.5cm]{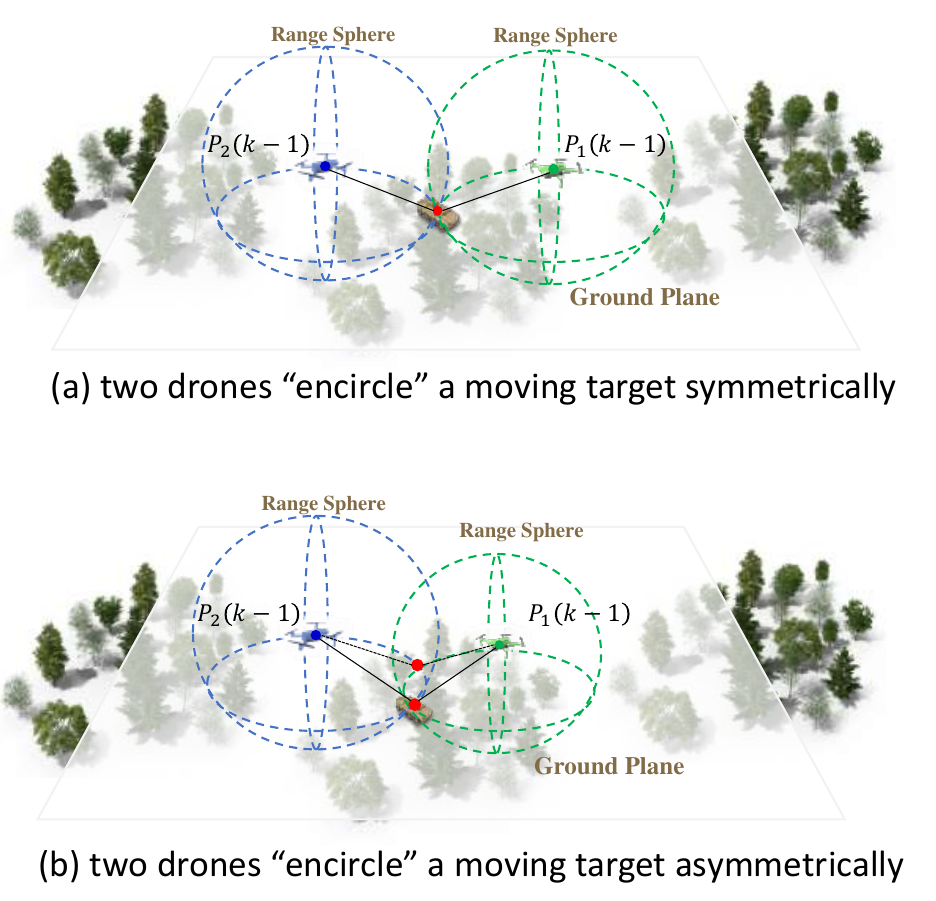}
  \vspace{-5pt}
 \caption{\footnotesize Two drones “encircle” a moving target.}
  \label{encircle_target}
   \vspace{-15pt}
\end{figure}

To ensure smooth encirclement of the target, a preset encirclement shape is designed for drone $i$ as \begin{equation}\label{eq4}.
\begin{split}
\bm{\mathcal{P}}=\rho
                  [\sin(\bar{\nu} k\pi),\cos(\bar{\nu} k\pi)]^\top ,
                  \end{split}
\end{equation}
where $\rho=\|\bm{\mathcal{P}}\|$ is the radius of the preset shape. $\bar{\nu}$ represents the frequency of circumnavigation and satisfies $\bar{\nu}=\frac{1}{\ell}$ with a positive integer $\ell\geq4$, which will be demonstrated in Lemma \ref{Uniform Observability}. In fact, $\bm{\mathcal{P}}$ is the desired relative position from drone to target. Denoting the relative position from drone $i$ to target $j$ as $\bp_{i,j}$, $\bm{\mathcal{P}}$ satisfies the following assumption.
\begin{assumption} \label{angle} (Angle to relative position)
To ensure drone $i$ encircles target $j$ while remaining opposite to one another drone $g, (g>i)$ (AS manner), the included angles between the preset shape and the relative position are set as $
\angle(\bm{\mathcal{P}},\bp_{i,j})=0^{\circ}$ and $\angle(\bm{\mathcal{P}},\bp_{g,j})=180^{\circ}$.
\end{assumption}
\begin{remark}
In this work, two drones symmetrically encircle the target in a specific shape, inspired by wolves' pincer attack strategy. This approach ensures effective target observation and control. As shown in Figure \ref{encircle_target}, when two drones symmetrically encircle the target, the target's position can be uniquely determined using just two distance measurements.
\end{remark}

\begin{figure}
\centering
  \includegraphics[width=7.1cm]{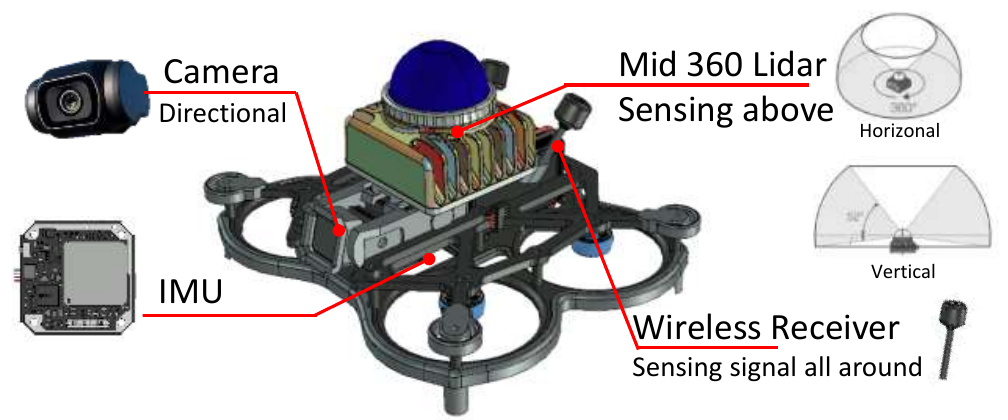}
  \vspace{-10pt}
 \caption{\footnotesize The onboard sensors configuration of drone.}
  \label{drone_configuration}
  \vspace{-15pt}
\end{figure}

\vspace{-0.5cm}
\subsection{Perception Measurements} \label{Distance_Measurement}
In this work, we assume prior knowledge of the targets, such as color and size, is available. Consequently, task drones are capable of recognizing targets, determining distances, and detecting nearby obstacles using onboard sensors. As shown in Figure \ref{drone_configuration}, the drone uses LiDAR to detect obstacles above the horizon, though LiDAR may struggle to detect small moving objects at a distance due to its limited resolution and range. To overcome this, the drone uses a gimbalized camera to estimate the target's distance through visual size cues \cite{vajgldist}, which can be fused with RSSI-based data for more robust range estimation.

To map the dimensions between the drone and the ground
target, we assume that all drone positions can be projected onto the ground plane to obtain a rough estimate of the distance measurement. The projection matrix is defined as:
\begin{equation*}
\begin{split}
F=\left[       \begin{array}{ccc}
                  1&0&0 \\
                  0&1&0 \\
                \end{array}
              \right].
\end{split}
\end{equation*}

Therefore, the distances from drone $i$ to target $j$ can be defined as $d_{i,j}=\|F\bx_i-\bs_j\|+\varepsilon_{i,j}, j \in \Phi_M$,
where $\varepsilon_{i,j} \in \mathbb{R}^1$ is stochastic measurement noises and satisfies zero-mean Gaussian distribution, i.e., $\E\{\varepsilon_{i,j}\}=0$ and $\Var\{\varepsilon_{i,j}\}=q_i$. Here, $q_i \in \mathbb{R}^+$ is a known constant. 
\begin{assumption} \label{mutiple measurements} 
There are $f$ independent identically distributed samplings of $d_{i,j}$ for drone $i$ within one sampling period $(k-1,k]$, i.e, $d_{i,j} \in \{d_{i,j}^{1},d_{i,j}^{2},\ldots, d_{i,j}^{f}\}$ for $f \in \mathbb{Z}^+$.
\end{assumption}
\begin{remark} \label{mutiple_measurements_remark} 
We assume the change in relative distance from the drone to the target within one sampling period is minimal. The drones continuously sample the target distance by fusing image size data or wireless signal strength at high frequencies (MHz and GHz), capturing $f$ discrete samples per interval. These multiple samplings differ from the multiple measurements mentioned in the Introduction, focusing on repeated sampling of a single distance measurement in the presence of noise.
\end{remark}

Based on Assumption \ref{Initial coordinate system matching}, the real-time position $\bx_{i}$ of drone $i$ is obtained by accumulating IMU data, without accounting for drift from noise or error accumulation.



\vspace{-0.1cm}
\subsection{Task Assignment} \label{Task Assignment}
In the initial phase, each drone assigns itself to the nearest target, ensuring two different drones choose a target within their measurement range. Due to communication and measurement limitations in Assumption \ref{Radius of communication and measurement}, each drone may not have access to all drone-to-target distances. Therefore, we modify the existing task assignment algorithm, CBAA from \cite{choi2009consensus}, to deploy two drones per target. The assignment process involves auction and consensus stages, designed as follows.

Firstly, in the auction process, the reward score of assigning target $j$ to drone $i$ is defined by $c_{i,j}(\tau)=c_{i,j,1}+c_{i,j,2}(\tau)$, where $\tau$ is the iteration counter that starts from zero, $c_{i,j,1}=\frac{1}{d_{i,j}}$ is the distance-based reward and $c_{i,j,2}(\tau)$ is the additional reward of target $j$. $c_{i,j,1}$ implies that the reward increases as the drone $i$ gets closer to the target $j$. $c_{i,j,2}(\tau)$ is continuously updated to reflect the current reward for the task. The initial value of $c_{i,j,2}(0)$ is set as zero. Furthermore, we need to store and update two vectors, i.e., the task vector $\mathcal{X}_i=[\tilde{x}_{i,j,g}]_{M\times N}, g\in \Phi_N, j \in \Phi_M$ and the score vector $\by_i(\tau)=[c_{i,j}(\tau)]_{1\times\si\{\Phi_{i,j}\}}$, where $\Phi_{i,j}$ and $\si\{\Phi_{i,j}\}$ represent a set of targets detected by drone $i$ and the number of targets, respectively. $\mathcal{X}_i$ is initialized as zero vector. Furthermore, when the $g$th column sum of $\mathcal{X}_i$ equals zero, i.e., $\sum_{j \in \Phi_M}\{\tilde{x}_{i,j,g}\}=0, g\in \Phi_{N}$, drone $i$ will select its target $j^{*}(\tau)$ in the $\tau$th iteration by $j^{*}(\tau)=\argmax_j\{c_{i,j}(\tau)\}, \forall j \in \Phi_{i,j}$. Then, we have $\tilde{x}_{i,j^*(\tau),i}=1$. The vector $\mathcal{X}_i$ can be further developed in the consensus process.

Secondly, in the consensus process, the drone $i$ will share $\mathcal{X}_{i}$ with its neighbors $\tilde{g}$. $\tilde{g}$ belongs to the set $\Phi_{i,\tilde{N}}\triangleq\{\tilde{g}| d_{i,\tilde{g}}\leq r_1\}$. The specific consensus process is shown in Algorithm \ref{The_task_sequence_execution}. The task vector $\mathcal{X}_i$ is updated based on the information received by drone $i$ (line 3). When the following three conditions are met simultaneously, drone $i$ will release the previously assigned tasks (line 5) and adjust the additional reward (line 6). The first condition is that the drone $i$ can observe multiple targets, i.e., $\si\{\Phi_{i,j}\}>1$. The second condition is that the number of drones assigned to the target $j^*(\tau)$ chosen by the drone $i$ is greater than two, i.e., $\sum_{g\in \Phi_{N}}\{\tilde{x}_{i,j^*(\tau),g}\}>2$. The third condition is that there exists a target observed by drone $i$ and chosen by no more than two drones, i.e., $\sum_{g\in \Phi_{N}}\{\tilde{x}_{i,h,g}\}\leq 2$, $\exists h\in \Phi_{i,j}$. A constant $(\tilde{\epsilon} > 1)$ is used to adjust additional rewards, allowing drone $i$ to assign a new target in the next round of the auction phase.

An iteration of the task assignment algorithm includes one run of the auction and consensus processes, which are asynchronous for each drone. If drone $i$ is assigned a task that remains unchanged during the consensus process, the auction is skipped in the next iteration, and only the consensus process runs. The task assignment for drone $i$ terminates in the initial phase when all observable targets have been selected by two different drones.

\begin{algorithm}
 \caption{The consensus process of drone $i$}
  \label{The_task_sequence_execution}
 \LinesNumbered
\KwIn{$\mathcal{X}_{i}$, $\by_{i}(\tau)$, $\mathcal{X}_{\tilde{g}}$, $\forall \tilde{g} \in \Phi_{i,\tilde{N}}$}
\KwOut{$c_{i,j,2}(\tau),\forall j\in \Phi_{i,j}$ and $\mathcal{X}_{i}$}
Send $\mathcal{X}_{i}$ to neighbors $\tilde{g}$, $\forall \tilde{g} \in \Phi_{i,\tilde{N}}$\;
Receive $\mathcal{X}_{\tilde{g}}$ from neighbors $\tilde{g}$, $\forall \tilde{g} \in \Phi_{i,\tilde{N}}$\;
$\tilde{x}_{i,j,g}=\max\{\tilde{x}_{\tilde{g},j,g}\}$, $\forall g\in \Phi_{N}$ and $\tilde{g} \in \Phi_{i,\tilde{N}}$\;
\If{$\si\{\Phi_{i,j}\}>1$, $\sum_{g\in \Phi_{N}}\{\tilde{x}_{i,j^*(\tau),g}\}>2$ and $\sum_{g\in \Phi_{N}}\{\tilde{x}_{i,h,g}\}\leq 2$, $\exists h\in \Phi_{i,j}$}{
$\tilde{x}_{i,j^*(\tau),i}=0$\;
$c_{i,h,2}(\tau)=\tilde{\epsilon}\{c_{i,j^*(\tau)}-c_{i,h}(\tau)\}$\;
}
\end{algorithm}

\vspace{-0.5cm}
\subsection{Problem Formulation} \label{problem}
\begin{figure}
\centering
  \includegraphics[width=8.5cm]{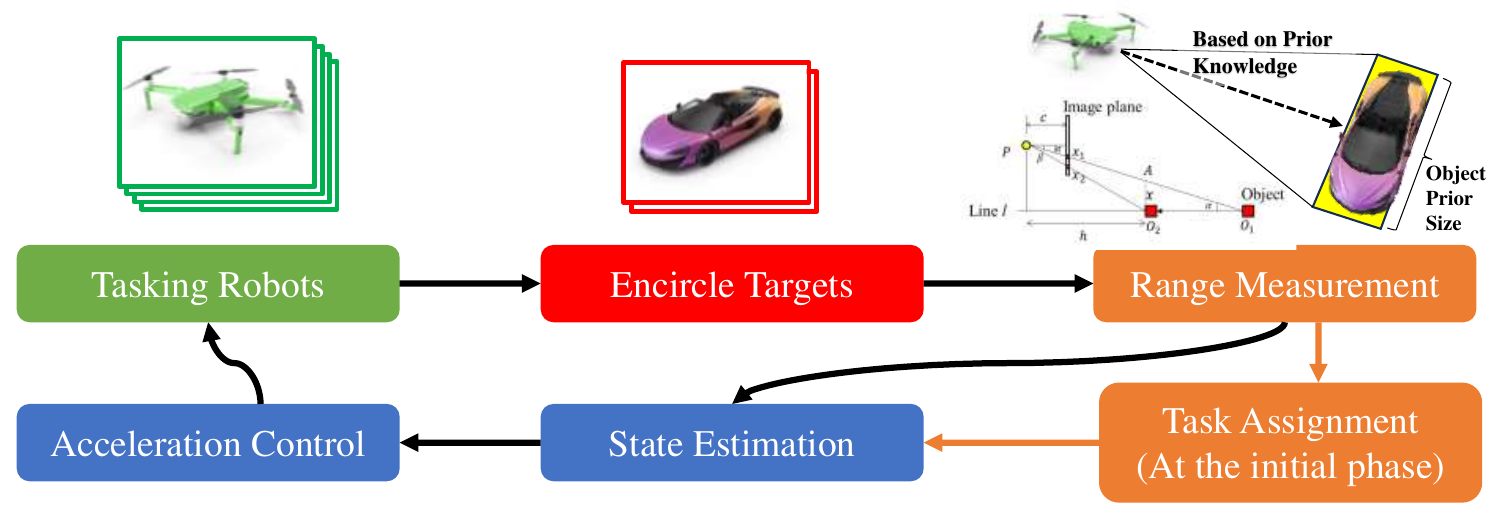}
  \vspace{-10pt}
 \caption{\footnotesize Architecture of multiple non-cooperative ground target encirclement strategy.}
 \vspace{-20pt}
  \label{whole_process}
\end{figure}

The architecture of our work is shown in Figure \ref{whole_process}, which is mainly composed of range measurement, task assignment, state estimation, and acceleration control. This article aims to design the estimator, defined as $\hat{\bm{\eta}}_j, j\in\Phi_M$, for each ground moving target $j$, and the acceleration controller $\bu_i$ for each tasking drone $i, i\in\Phi_N$ such that the tasking drones in \eqref{eq2} can encircle all ground moving targets in \eqref{eq3} while avoiding all collisions. The objectives are described as follows:
\begin{enumerate}[(i)]
\item \label{estimation} (Moving target state estimation)  
Under the process and measurement noises, the state of moving target $j$ is estimated such that the estimation error, denoted as $\be_{j}=\bm{\eta}_j-\hat{\bm{\eta}}_j$, is bounded in mean square, i.e.,
$\E\{\|\be_{j}\|^2\}\leq \delta_1, \forall j\in\Phi_M$,
where $\delta_1 \in \mathbb{R}$ is non-negative error bound.
\item (Target encirclement) \label{encirclement}
The target encirclement part consists of the following two objectives: 
\begin{enumerate}[(1)]
\item  Robot $i, i\in\Phi_N$ can avoid all obstacles, other drones, and targets.
\item  Under Assumption \ref{angle}, drones $i$ and $g$ in one group can encircle the target $j$ along a specific shape $\bm{\mathcal{P}}$ and are positioned symmetrically on either side of the target $j$ to enable the position measurements of the target $j$, i.e., $\E\{\|\bp_{i,j}+\bm{\mathcal{P}}\|^2\}\leq\delta_{3,1}$ and $\E\{\|\bp_{g,j}-\bm{\mathcal{P}}\|^2\}\leq\delta_{3,2}$, simplified as $\E\{\|\bp_{i,j}+\bp_{g, j}\|^2\}\leq\delta_3,\forall i,g\in\Phi_N,i\neq g$,
where $\bp_{i,j}=F\bx_i-\bs_j$ and $\bp_{g,j}=F\bx_g-\bs_j$ are the relative positions, $\delta_{3} \leq 2\delta_{3,1}+2\delta_{3,2} $ is non-negative error bound.
\end{enumerate}
\end{enumerate}

\vspace{-0.2cm}
\section{Position estimation and control law}
In this section, all the target states are estimated firstly based on the task assignment algorithm and the noisy distance information. Then, a new acceleration controller is designed to ensure that the drones can quickly encircle the targets in a multi-obstacle environment.

\vspace{-0.2cm}
\subsection{Target State Estimator}
Based on the task assignment algorithm in Sec. \ref{Task Assignment}, all drones can be divided into $M$ groups in the initial phase. We assume that each group of drones, denoted by $i$ and $g$, can collaboratively perceive targets, marked by the sets $\Phi_{i,g,\bar{M}}$.

Note that the distance measurements $d_{i,j}$ and $d_{g,j}$, and the positions $\bx_i$ and $\bx_g$ of drones $i$ and $g$ are available and define $\theta_{i,g,j}\triangleq d_{i,j}^{2}-d_{g,j}^{2}-\bx_{i}^\top F^\top F\bx_{i}+\bx_{g}^\top F^\top F\bx_{g}$ as a measurement of the target $j, j\in \Phi_{i,g,\bar{M}}$. Then, considering $
d_{i,j}=\|F\bx_i-\bs_j\|+\varepsilon_{i,j}$ with the noise $\varepsilon_{i,j}$ and $\|F\bx_i-\bs_j\|^2=\bx_i^\top F^\top F\bx_i-2\bx_i^\top F^\top\bs_j+\bs_j^T\bs_j$, $\theta_{i,g,j}$ can be rewritten as
\begin{equation}\label{eq5}
\begin{split}
\theta_{i,g,j}=&C_{i,g}\bm{\eta}_j+\bar{\varepsilon}_{i,g,j},
\end{split}
\end{equation}
where $C_{i,g}=[-2\bp_{i,g}^{\top}F^{\top},0_{1\times 2}]$  with the relative position $\bp_{i,g}$. $\bar{\varepsilon}_{i,g,j}=\varepsilon_{i,j}^2-\varepsilon_{g,j}^2+2(\|F\bx_i-\bs_j\|)\varepsilon_{i,j}-2(\|F\bx_g-\bs_j\|)\varepsilon_{g,j}$ is the output noise, which has the mean $\E\{\bar{\varepsilon}_{i,g,j}\}=q_{i}-q_{g}$ since $\E\{\varepsilon_{i,j}^2\}=q_{i}$ and $\E\{\varepsilon_{g,j}^2\}=q_{g}$. Furthermore, define the variance of $\bar{\varepsilon}_{i,g,j}$ as $\Upsilon_{i,g,j}\triangleq\Var\{\bar{\varepsilon}_{i,g,j}\}$. Based on $\Upsilon_{i,g,j}=\E\{\bar{\varepsilon}_{i,g,j}^2\}-\E\{\bar{\varepsilon}_{i,g,j}\}^2$, we have 
\begin{equation}\label{eq6}
\begin{split}
\Upsilon_{i,g,j}
=&\E\{\varepsilon_{i,j}^4\}+4\E\{\|F\bx_i-\bs_j\|^2\varepsilon_{i,j}^2\}\\
&+\E\{\varepsilon_{g,j}^4\}+4\E\{\|F\bx_g-\bs_j\|^2\varepsilon_{g,j}^2\}\\
&-2\E\{\varepsilon_{i,j}^2\}\E\{\varepsilon_{g,j}^2\}-\E\{\bar{\varepsilon}_{i,g,j}\}^2.
\end{split}
\end{equation}


Substituting $\E\{\varepsilon_{i,j}^4\}=3q_i^2$ and $\E\{\varepsilon_{g,j}^4\}=3q_g^2$ into \eqref{eq6}, the variance $\Upsilon_{i,g,j}$ can be estimated as
\begin{equation}\label{eq7}
\begin{split}
\hat{\Upsilon}_{i,g,j}=&2q_{i}^2+2q_{g}^2+4(\frac{1}{f}\sum_{\varrho=1}^{f}(d_{i,j}^{\varrho})^2-q_i)q_{i}\\
&+4(\frac{1}{f}\sum_{\varrho=1}^{f}(d_{g,j}^{\varrho})^2-q_g)q_{g},
\end{split}
\end{equation}
where $\frac{1}{f}\sum_{\varrho=1}^{f}(d_{i,j}^{\varrho})^2-q_i\approx\E\{\|F\bx_i-\bs_j\|^2\}$ for $\E\{d_{i,j}^2\}\approx \frac{1}{f}\sum_{\varrho=1}^{f}(d_{i,j}^{\varrho})^2$ based on Assumption \ref{mutiple measurements}. Considering $\|F\bx_i-\bs_j\|^2$ always being non-negative, we have $d_{i,j}^2-q_i\geq0$.

By recalling the target model \eqref{eq2} and the measurement output \eqref{eq5}, and based on the Kalman filtering method in \cite{li2016new,reif1999stochastic}, the distance-only-based target state estimator (DTSE) can be designed as
\begin{subequations}\label{eq8}
\begin{align}
\hat{\bm{\eta}}_{j}=&A_2\hat{\bm{\eta}}_j^{(-)}+K_j(\theta_{i,g,j}-\hat{\theta}_{i,g,j}),\label{eq8-1}\\
\hat{\theta}_{i,g,j}=&C_{i,g}A_2\hat{\bm{\eta}}_j^{(-)}+q_{i}-q_{g},\label{eq8-2}\\
\bm{\zeta}_j=&(I_{4\times 4}-K_jC_{i,g})\bm{\zeta}_j^{(k|k-1)}\label{eq8-3},\\
\bm{\zeta}_j^{(k|k-1)}=&A_2\bm{\zeta}_j^{(-)}A_2^\top +B_2Q_{j} B_2^\top \label{eq8-4},\\
K_j=&\bm{\zeta}_j^{(k|k-1)}C_{i,g}^{\top}(C_{i,g}\bm{\zeta}_j^{(k|k-1)} C_{i,g}^{\top}+\hat{\Upsilon}_{i,g,j})^{-1}\label{eq8-5}.
\end{align}
\end{subequations}
In formula \eqref{eq8-1}, $K_j \in \mathbb{R}^{4\times 1}$ is the estimator gain. $\hat{\theta}_{i,g,j}$ represents the estimated measurement output of target $j$. In formula \eqref{eq8-2}, the term $A_2\hat{\bm{\eta}}_j^{(-)}\triangleq \hat{\bm{\eta}}_j^{(k|k-1)}$ is the predicted state according to the model \eqref{eq2}. Since the measurement output noise $\bar{\varepsilon}_{i,g,j}$ in \eqref{eq5} is unknown, we directly use the mean $q_{i}-q_{g}$ to compensate the output $\hat{\theta}_{i,g,j}$. In formula \eqref{eq8-3}, $\bm{\bm{\zeta}}_j=\E\{\be_j\be_j^{\top}\}$ is the estimation error variance, where $\be_j$ is defined in Sec. \ref{problem} (i). $\bm{\zeta}_j^{(k|k-1)}$ is the predicted error variance, and the initial variance $\bm{\zeta}_j^{(0)}$ is a given positive definite matrix.
Therefore, the estimated position and velocity of the target can be obtained as
\begin{equation}\label{eq9}
\begin{split}
\hat{\bs}_j=\check{I}_{2\times 2}\hat{\bm{\eta}}_{j},~\hat{\nu}_j=\hat{I}_{2\times 2}\hat{\bm{\eta}}_{j},
\end{split}
\end{equation}
where $\check{I}_{2\times 2}=[I_{2\times 2},0_{2\times 2}]$ and $\hat{I}_{2\times 2}=[0_{2\times 2},I_{2\times 2}]$.

By combining \eqref{eq8-1} and \eqref{eq8-2}, the dynamics of the predicted state $\hat{\bm{\eta}}_j^{(k+1|k)}$ can be obtained as
\begin{subequations}\label{eq10}
\begin{align}
\hat{\bm{\eta}}_j^{(k+1|k)}=&A_2\hat{\bm{\eta}}_j^{(k|k-1)}+A_2K_j(\theta_{i,g,j}-\hat{\theta}_{i,g,j}^{(k+1|k)}),\label{eq10-1}\\
\hat{\theta}_{i,g,j}^{(k+1|k)}=&C_{i,g}\hat{\bm{\eta}}_j^{(k+1|k)}+q_{i}-q_{g}.\label{eq10-2}
\end{align}
\end{subequations}

We denote the predicted state error as $\be_{j}^{(k|k-1)}=\bm{\eta}_j-\hat{\bm{\eta}}_j^{(k|k-1)}$. Furthermore, the dynamic of the predicted state error can be deduced as
\begin{equation}\label{eq10-3}
\begin{split}
\be_j^{(k+1|k)}=&A_2(I_{4\times 4}-K_jC_{i,g})\be_j^{(k|k-1)}+B_2\bm{\omega}_j\\
&-A_2K_j(\bar{\varepsilon}_{i,g,j}-(q_{i}-q_{g})).
\end{split}
\end{equation}

\subsection{Encirclement Accelerate Controller}
All drones must navigate around obstacles while encircling targets using pseudo-forces, including attraction to the target, mutual interaction between drones, and repulsion from obstacles and other targets, as in \cite{li2022three}.


Firstly, the attractive pseudo-force is designed to ensure the encirclement of the target. In the initial phase, drones $i$ and $g$ are assigned to encircle target $j$, with forces in opposite directions. Then, the attractive pseudo-forces are described by:
\begin{equation*}
\begin{split}
\Gamma_{at,i}=&-\frac{2}{t^2}F^\top\{ \hat{\bp}_{i,j}-\bm{\mathcal{P}}-t\hat{\nu}_j\},\\
\Gamma_{at,g}=&-\frac{2}{t^2}F^\top \{  \hat{\bp}_{g,j}+\bm{\mathcal{P}}-t\hat{\nu}_j\},
\end{split}
\end{equation*}
where the terms $\hat{\bp}_{i,j}=F\bx_i-\hat{\bs}_{j}$ and $\hat{\bp}_{g,j}=F\bx_g-\hat{\bs}_{j}$ are the estimated relative positions.
When only the attractive field is considered, the acceleration controller $\bu_i$ for drone $i$ is designed as follows:
\begin{equation}\label{eq14}
\begin{split}
\bu_i=\Gamma_{at,i}-\frac{2}{t}\bm{v}_i.
\end{split}
\end{equation}

Denote the AS-based encirclement error as $\bar{\be}_j=(F\bx_i-\bs_j)+(F\bx_g-\bs_j)$. Furthermore, based on the state models in \eqref{eq2} and \eqref{eq3}, and the acceleration controller in \eqref{eq14}, the dynamic of $\bar{\be}_j$ in this case is derived as
\begin{equation}\label{eq15}
\begin{split}
\bar{\be}_j^{(+)}=&-2\check{I}_{2\times 2}A_2\be_j-t^2\bm{\omega}_j.
\end{split}
\end{equation}

Secondly, the mutual interaction and repulsive pseudo-forces minimize collision risks and protect the drones. The safety radius $\tilde{r} = \tilde{a} + \tilde{b}$ is defined by the drone's radius $\tilde{a}$ and the minimum safe distance $\tilde{b}$. The pseudo-force action radius, $\bar{r}_{j}$, depends on the target's real-time velocity, ensuring effective tracking at high speeds. It is given by $\bar{r}_{j} = \tilde{r} + \Delta r + |t\hat{\bm{\nu}}_j|$, where $\Delta r$ is a positive constant.

Recalling $\Phi_{i,\tilde{N}}$ defined in Sec. \ref{Task Assignment}, the mutual interaction pseudo-force $\Gamma_{in,i}$ of the drone $i$, $i\in \Phi_N$ is described by the following function,
\begin{equation*}
\begin{split}
\Gamma_{in,i}=&\sum_{\tilde{g}\in\Phi_{i,\tilde{N}}}\frac{-2\gamma_1\tilde{r}\bp_{i,\tilde{g}}\step\{2\bar{r}_{j}-d_{i,\tilde{g}}\}\step\{d_{i,\tilde{g}}-2\tilde{r}\}}{(2\tilde{r}-d_{i,\tilde{g}})d^2_{i,\tilde{g}}},
\end{split}
\end{equation*}
where $\bp_{i,\tilde{g}}=\bx_i-\bx_{\tilde{g}}$ and $d_{i,\tilde{g}}=\|\bp_{i,\tilde{g}}\|$. $\gamma_1$ is a positive coefficient. $\step$ is the step function, i.e., $\step\{2\bar{r}_{j}-d_{i,\tilde{g}}\}=1$ if $2\bar{r}_{j}\geq d_{i,\tilde{g}}$, otherwise $\step\{2\bar{r}_{j}-d_{i,\tilde{g}}\}=0$. 

Let all targets and obstacles that can be positioned by drone $i$ belong to the set $\Phi_{i,\bar{M},\bar{O}}\triangleq\Phi_{i,g,\bar{M}}\cup\Phi_{i,\bar{O}}$, where the set $\Phi_{i,\bar{O}}$ represents the obstacles detected by drone $i$. The repulsive pseudo-force $\Gamma_{re,i}$ of the drone $i$, $i\in \Phi_N$ based on the state estimators is described as
\begin{equation*}
\begin{split}
\Gamma_{re,i}=&
\sum_{\iota\in \Phi_{i,\bar{M},\bar{O}}}\frac{-\gamma_2\tilde{r}\hat{\bp}_{i,\iota}\step\{\bar{r}_{j}-\hat{d}_{i,\iota}\}\step\{\hat{d}_{i,\iota}-\tilde{r}\}}{(\tilde{r}-\hat{d}_{i,\iota})d^2_{i,\iota}},
\end{split}
\end{equation*}
where $\hat{\bp}_{i,\iota}=\bx_i-\bo_\iota$ for $\iota \in \Phi_{i,\bar{O}}$, $\hat{\bp}_{i,\iota}=F\bx_i-\hat{\bs}_\iota$ for $\iota \in \Phi_{i,g,\bar{M}}$ and $\hat{d}_{i,\iota}=\|\hat{\bp}_{i,\iota}\|$. $\gamma_2$ is a positive coefficient. 

Consider a situation where task drone $i$ is positioned between obstacles $\iota \in \Phi_{i,\bar{M},\bar{O}}$ and the encircled target $j$, or between neighbor drone $\tilde{g} \in \Phi_{i,\tilde{N}}$ and target $j$. If drone $i$ is too close to obstacles $\iota$ or neighbor $\tilde{g}$, the pseudo-forces $\Gamma_{in,i}$ and $\Gamma_{re,i}$ may become infinite and align with $\Gamma_{at,i}$, risking a collision with target $j$. To prevent this, when the angles between the pseudo-forces $\Gamma_{in,i}$, $\Gamma_{re,i}$, and $\Gamma_{at,i}$ are within $[0^{\circ}, 90^{\circ}]$, an upper limit $\epsilon$ is imposed on the pseudo-forces. The mutual interaction and repulsive pseudo-forces are then redesigned as follows:

For all $
\angle(\Gamma_{in,i},\Gamma_{at,i})\cap \angle(\Gamma_{re,i},\Gamma_{in,i})\in [0^{\circ},90^{\circ}] $ or $
\angle(\Gamma_{in,i},\Gamma_{at,i})\in [0^{\circ},90^{\circ}] \cap (\Gamma_{re,i}=0)$, then
\begin{equation*}
\begin{split}
\bar{\Gamma}_{in,i}=&\frac{\epsilon\Gamma_{in,i}}{\max\{\epsilon,\|\Gamma_{in,i}\|\}}.
\end{split}
\end{equation*}

For all $
\angle(\Gamma_{re,i},\Gamma_{at,i})\cap \angle(\Gamma_{re,i},\Gamma_{in,i})\in [0^{\circ},90^{\circ}]$ or $
\angle(\Gamma_{re,i},\Gamma_{at,i})\in [0^{\circ},90^{\circ}]\cap (\Gamma_{in,i}=0)$, then
\begin{equation*}
\begin{split}
\bar{\Gamma}_{re,i}=&\frac{\epsilon\Gamma_{re,i}}{\max\{\epsilon,\|\Gamma_{re,i}\|\}}.
\end{split}
\end{equation*}

Therefore, the resultant pseudo-forces of the drones $i$, $i\in \Phi_N$ can be obtained as
\begin{equation}\label{eq16}
\begin{split}
\Gamma_{i}=&\Gamma_{at,i}+\bar{\Gamma}_{in,i}+\bar{\Gamma}_{re,i}.
\end{split}
\end{equation}

Furthermore, a novel acceleration controller $\bu_i, i\in \Phi_N$ of drone $i$ is designed as
\begin{equation}\label{eq17}
\begin{split}
\bu_i=\Gamma_{i}-\frac{2}{t}\bm{v}_i.
\end{split}
\end{equation}

\begin{remark}
Unlike the navigation or path planning methods in \cite{pan2021improved, yan2012multilevel}, we incorporate the AS technique \cite{liu2023multiple, liu2023moving} into the attractive pseudo-force to position two drones on opposite sides of the target, maximizing sensor coverage. The attraction force remains active even when both the drone and target are stationary, avoiding local optima where the sum of pseudo-forces is zero. Additionally, an upper bound on the repulsive pseudo-force prevents collisions with the target, making it suitable for dynamic environments with multiple moving objects.
\end{remark}

\vspace{-0.2cm}
\section{Convergence analysis}
This section first presents assumptions and lemmas to establish system stability, followed by convergence analysis of estimation errors to ensure accurate state estimation in Theorem 1. It then analyzes AS-based encirclement error convergence under attractive pseudo-forces in Theorem 2 and discusses obstacle avoidance effectiveness under the acceleration controller in Theorem 3.



\begin{assumption} \label{limit} There exist positive real constants  $\hat{q}$, $\check{q}$, $\hat{r}$ and $\check{r}$ such that the noise variances satisfy the following conditions,
\begin{subequations}\label{eq18}
\begin{align}
&\hat{q}I_{2\times 2}\leq Q_j\leq \check{q}I_{2\times 2}, \label{eq18-2}\\
&\hat{r}\leq\Upsilon_{i,g,j}\leq \check{r}. \label{eq18-3}
\end{align}
\end{subequations}
\end{assumption}

\begin{remark} 
According to Remark \ref{input noise variance}, we assume the target's maximum acceleration is known, allowing us to estimate the bounds of the variance $Q_j$. Given formulas \eqref{eq6} and \eqref{eq7}, and the known positive variances $q_i$ and $q_g$, along with the non-negative means $\E\{\|F\bx_i-\bs_j\|^2\}$ and $\E\{\|F\bx_g-\bs_j\|^2\}$, we can assume that $\Upsilon_{i,g,j}$ has strict positive definite bounds.
\end{remark}

\begin{assumption} \label{without_repulsive_force} 
Assume that there exists a sufficiently large positive integer $m_1$ such that for all $k\geq m_1$ and within the time interval $[k-m_1,k]$, there are always four or more consecutive time instants where the sums of the repulsive and internal forces acting on drones $i$ and $g$ are equal to zero, i.e., $\bar{\Gamma}_{in, i}+\bar{\Gamma}_{re, i}=0$ and $\bar{\Gamma}_{in,g}+\bar{\Gamma}_{re,g}=0$.
\end{assumption}

\begin{remark}
According to the expressions for $\bar{\Gamma}_{in, i}$, $\bar{\Gamma}_{re, i}$, $\bar{\Gamma}_{in,g}$, and $\bar{\Gamma}_{re,g}$, when obstacles or other drones enter the region with radius $\bar{r}_j$ around drones $i$ and $g$, the repulsive forces increase significantly. The drone then moves away from obstacles to reduce the repulsive force to zero. Thus, it is reasonable to assume that within the time interval $[k-m_1, k]$, there are at least four consecutive instants when repulsive and interaction forces are inactive.
\end{remark}


\begin{lemma}(Uniform observability)\label{Uniform Observability}
Under Assumption \ref{limit} and Assumption \ref{without_repulsive_force}, the state model of target $j$ in \eqref{eq3} with output equation \eqref{eq5} is uniformly observable, implying that the observability Gramin matrix $\mathcal{O}_1$ satisfies $\Lambda_1I_{4\times4}\leq \mathcal{O}_1\leq\Lambda_2I_{4\times4}$, where $\Lambda_1$ and $\Lambda_2$ are positive constants.
\end{lemma}
\textbf{Proof.}
See Appendix A.

\begin{lemma}(Controllability) \label{Controllable}
Given the sampling period $t>0$ and under Assumption \ref{limit}, the state model of target $j$ in \eqref{eq3} is controllable, indicating that the controllability Gramin matrix $\mathcal{H}_2\hat{Q}_j\mathcal{H}_2^\top$  satisfies $\Lambda_3I_{4\times4}\leq\mathcal{H}_2\hat{Q}_j\mathcal{H}_2^\top\leq\Lambda_4I_{4\times4}$, where $\Lambda_3$ and $\Lambda_4$ are positive constants, 
the diagonal matrix $\hat{Q}_j=\diag\{Q_j,\ldots,Q_j\} \in \mathbb{R}^{2(m_1+1) \times 2(m_1+1)}$, and the controllability matrix 
\begin{equation*}
\begin{split}
\mathcal{H}_2=
\left[
                   \begin{array}{cccc}
                     \frac{1}{2}t^2&\frac{3}{2}t^2&\cdots&\frac{2m_1+1}{2}t^2\\
                      t &t&\cdots&t\\
                   \end{array}
                 \right] \otimes I_{2\times 2}.
\end{split}
\end{equation*}
\end{lemma}

\begin{lemma} \label{variance bound}
Under Assumption \ref{limit}, Lemma \ref{Uniform Observability} and Lemma \ref{Controllable}, there always exists positive bounds $\check{\zeta}_j$ and $\hat{\zeta}_j$ such that the estimation error variance $\bm{\zeta}_j$ satisfies
\begin{subequations}\label{eq2-41}
\begin{align}
&\hat{\zeta}_jI_{4\times4}\leq\bm{\zeta}_j^{-1}\leq\check{\zeta}_jI_{4\times4},\label{eq2-41-1}
\end{align}
\end{subequations}
for all $k>0$.
\end{lemma}

\textbf{Proof.}
See Appendix B.


\begin{theorem} \label{estimation error convergence}
Under Assumption \ref{limit} and Lemma \ref{variance bound}, the estimation error $\be_j$ of the target is bounded in mean square.
\end{theorem}

\textbf{Proof.}
Firstly, to prove the convergence of the predicted state error  $\be_j^{(k|k-1)}$, $j \in\Phi_M$ in \eqref{eq10-3}, the following Lyapunov function will be chosen,
\begin{equation*}
\begin{split}
\V_{1,j}^{(k|k-1)}=(\be_j^{(k|k-1)})^{\top}(\bm{\zeta}_j^{(k|k-1)})^{-1}\be_j^{(k|k-1)}.
\end{split}
\end{equation*}

Considering \eqref{eq2-41-1} in Lemma \ref{variance bound}, \eqref{eq8-4}  and Assumption \ref{limit}, we have
\begin{equation*}
\begin{split}
0_{4 \times4}<(\bar{a}\hat{\bm{\zeta}}_j^{-1}+\bar{b}\check{q})^{-1}I_{4 \times4}\leq(\bm{\zeta}_j^{(k|k-1)})^{-1}\leq\check{\bm{\zeta}}_j(\underline{a})^{-1}I_{4 \times4},
\end{split}
\end{equation*}
and then we arrive at
\begin{equation*}
\begin{split}
(\bar{a}\hat{\bm{\zeta}}_j^{-1}+\bar{b}\check{q})^{-1}\|\be_j^{(k|k-1)})\|^2\leq &\V_{1,j}^{(k|k-1)}\\
&
\leq\check{\bm{\zeta}}_j(\underline{a})^{-1}\|\be_j^{(k|k-1)})\|^2.
\end{split}
\end{equation*}

Furthermore, the difference of the Lyapunov function in mean square can be derived as
\begin{equation*}
\begin{split}
\E\{\triangle \V_{1,j}^{(k|k-1)}\}
=&\E\big\{(\be_j^{(k|k-1)})^\top (I_{4\times 4}-K_jC_{i,g})^\top A_2^\top \\
&\times(\bm{\zeta}_j^{(k+1|k)})^{-1}A_2(I_{4\times 4}-K_j C_{i,g})\be_j^{(k|k-1)}\\
&-(\be_j^{(k|k-1)})^{\top}(\bm{\zeta}_j^{(k|k-1)})^{-1}\be_j^{(k|k-1)}\big\}\\
&+\E\{\bm{\omega}_j^\top B_2^\top(\bm{\zeta}_j^{(k+1|k)})^{-1}B_2\bm{\omega}_j\}\\
&+\Upsilon_{i,g,j}K_j^{\top}A_2^{\top}(\bm{\zeta}_j^{(k+1|k)})^{-1}A_2K_j.
\end{split}
\end{equation*}

Recalling formula \eqref{eq8-4} again, we have $A_2^\top(\bm{\zeta}_j^{(k+1|k)})^{-1}A_2 \leq \bm{\zeta}_j^{-1}$,
and following formula \eqref{eq8-3}, we can further have
\begin{equation*}
\begin{split}
(I_{4\times 4}-K_jC_{i,g})^\top A_2^\top&(\bm{\zeta}_j^{(k+1|k)})^{-1}A_2(I_{4\times 4}-K_j C_{i,g})\\
&\leq (I_{4\times 4}-K_jC_{i,g})^\top(\bm{\zeta}_j^{(k|k-1)})^{-1}.
\end{split}
\end{equation*}

Let $G^{(k|k-1)}\triangleq C_{i,g}\bm{\zeta}_j^{(k|k-1)} C_{i,g}^{\top}+\hat{\Upsilon}_{i,g,j}$. Considering $C_{ij}^{\top}C_{ij}\leq 4\check{c}I_{4 \times 4}$ and Assumption \ref{limit}, we have $G^{(k|k-1)}\leq\big(4\check{c}\hat{r}^{-1}(\bar{a}\hat{\bm{\zeta}}_j^{-1}+\bar{b}\check{q})+1\big)\hat{\Upsilon}_{i,g,j}\triangleq \check{G}\hat{\Upsilon}_{i,g,j}$, where $\check{G}>1$. From \eqref{eq8-5}, we derive $(K_jC_{i,g})^\top(\bm{\zeta}_j^{(k|k-1)})^{-1}=C_{i,g}^\top(G^{(k|k-1)})^{-1}C_{i,g}$. Therefore, we have $(I_{4\times 4}-K_jC_{i,g})^\top(\bm{\zeta}_j^{(k|k-1)})^{-1} \leq(\bm{\zeta}_j^{(k|k-1)})^{-1}-C_{i,g}^\top(\check{G}\hat{\Upsilon}_{i,g,j})^{-1}C_{i,g}$.

Moreover, using formula \eqref{eq8-3}, the estimator gain can be rewritten as $K_j=\bm{\zeta}_jC_{i,g}^\top \hat{\Upsilon}_{i,g,j}^{-1}$. Based on Assumption \ref{limit} and Lemma \ref{variance bound}, we have $\|K_j\|\leq \check{K}$, where $\check{K}=2\sqrt{\check{c}}\hat{\zeta}_j^{-1}\hat{r}^{-1}$. Then, since $\E\{(\be_j^{(k|k-1)})^\top\be_j^{(k|k-1)}\}\geq\E\{\be_j^{(k|k-1)}\}^\top\E\{\be_j^{(k|k-1)}\}$ and $C_{i,g}^\top (\Upsilon_{i,g,j})^{-1} C_{i,g}\geq0_{4\times4}$, $\E\{\triangle \V_{1,j}^{(k|k-1)}\}$ can be re-derived as
\begin{equation*}
\begin{split}
\E\{\triangle \V_{1,j}^{(k|k-1)}\}
\leq&- \check{G}^{-1}\E\{\be_j^{(k|k-1)}\}^\top C_{i,g}^\top (\Upsilon_{i,g,j})^{-1}\\
&\times C_{i,g}\E\{\be_j^{(k|k-1)}\}+\tilde{\delta}_1,
\end{split}
\end{equation*}
where $\tilde{\delta}_1=\{\bar{b}\check{q}+\check{K}^2\check{r}\bar{a}\}\underline{a}^{-1}\check{\bm{\zeta}}_j$.

Defined $\bar{A}\triangleq A_2(I_{4\times 4}-K_jC_{i,g})$, and from formula \eqref{eq10-3}, we have $\E\{\be_j^{(k+1|k)}\}=\bar{A}\be_j^{(k|k-1)}$. Considering $I_{4\times 4}-K_jC_{i,g}=\bm{\zeta}_j(\bm{\zeta}_j^{(k|k-1)})^{-1}>0_{4\times4}$, similar to the results in Lemma \ref{Uniform Observability}, we can obtain that $\bar{\mathcal{O}}\triangleq\sum^{k}_{\varrho=k-m_1}(\bar{\mathcal{A}}^{\varrho})^{\top}(C_{i,g}^{(\varrho)})^{\top}(\Upsilon_{i,g,j}^{(\varrho)})^{-1}C_{i,g}^{(\varrho)}\bar{\mathcal{A}}^{\varrho}$ has strictly positive upper and lower bounds, where $\bar{\mathcal{A}}^{k}=I_{4\times 4}$, $\bar{\mathcal{A}}^{\varrho}=\bar{A}^{k-\varrho}$. Therefore, we can conclude that
\begin{equation*}
\begin{split}
\E&\{\V_{1,j}^{(k|k-1)}-\V_{1,j}^{(k-m_1|k-m_1-1)}\}\\
\leq &- \check{G}^{-1}(\be_j^{(k-m_1|k-m_1-1)})^\top \bar{\mathcal{O}}\be_j^{(k-m_1|k-m_1-1)}+m_1\tilde{\delta}_1\\
\leq &-\lambda_{\min}\{\bar{\mathcal{O}}\}\check{G}^{-1}\underline{a}\check{\bm{\zeta}}_j^{-1}\V_{1,j}^{(k-m_1|k-m_1-1)}+m_1\tilde{\delta}_1\\
\leq &-\varpi\V_{1,j}^{(k-m_1|k-m_1-1)}+m_1\tilde{\delta}_1,
\end{split}
\end{equation*}
where $\varpi$ is a positive constant that satisfies $0<\varpi\leq1$ and $\varpi\leq\lambda_{\min}\{\bar{\mathcal{O}}\}\check{G}^{-1}\underline{a}\check{\bm{\zeta}}_j^{-1}$.

Based on [Lemma 2.1, \cite{reif1999stochastic}], the predicted state error  $\be_j^{(k|k-1)}$ in \eqref{eq10-3} is regarded as exponentially bounded in mean square, which leads to $\E\{\|\be_j^{(k|k-1)}\|^2\}\leq \bar{\delta}_1$ with the positive bound $\bar{\delta}_1$.

Secondly, we aim to prove estimation error $\be_j$ is bounded in mean square. For $\be_j^{(k+1|k)}=A_2\be_j+B_2\bm{\omega}_j$, we have
\begin{equation*}
\begin{split}
\E\{\|\be_j\|^2\}\leq &\frac{1}{\lambda_{\min}\{A_2^\top A_2\}}(\E\{\|\be_j^{(k+1|k)}\|^2\}\\
&-\lambda_{\min}\{B_2^\top B_2\}\E\{\|\bm{\omega}_j\|^2\}).
\end{split}
\end{equation*}
Therefore, the objective \eqref{estimation}) in Sec. \ref{problem} can be achieved with $\delta_1=\bar{\delta}_1\underline{a}^{-1}$.

\begin{theorem}
If only the attractive pseudo-force $\Gamma_{at,i}$ is applied, the target $j$ can be encircled by drone $i$ and drone $g$ in the AS manner.
\end{theorem}
\textbf{Proof.}
The Lyapunov function is chosen as $\V_{2}=\|\bar{\be}_j\|^2$. Furthermore, based on formula \eqref{eq15}, and the condition $\E\{\bm{\omega}_j\}=0$, the dynamic of the Lyapunov function in mean square can be obtained as
\begin{equation}\label{eq2-63}
\begin{split}
\E\{\V_{2}^{(+)}\}=&\E\{(2\check{I}_{2\times 2}A_2\be_j+t^2\bm{\omega}_j)^\top(2\check{I}_{2\times 2}A_2\be_j+t^2\bm{\omega}_j)\}\\
=&4\E\{\be_j^\top A_2^\top\check{I}_{2\times 2}^\top\check{I}_{2\times 2}A_2\be_j\}+t^4\E\{\|\bm{\omega}_j\|^2\}.
\end{split}
\end{equation}

Furthermore, considering $\E\{\|\be_{j}\|^2\}\leq \delta_1$ as proven in Theorem \ref{estimation error convergence},  $\E\{\|\bm{\omega}_j\|^2\}=\trace\{\Var\{\bm{\omega}_j\}\} \leq2\check{q}$ in Assumption \ref{limit}, and $\lambda_{\max}\{A_2A_2^\top\}=\lambda_{\max}\{A_3^\top A_3\}=\bar{a}$, we have
\begin{equation}\label{eq2-64}
\begin{split}
\E\{\V_{2}^{(+)}\}\leq4\bar{a}\delta_1+2t^4\check{q},
\end{split}
\end{equation}
for all $k>0$. Therefore, the encirclement error in mean square is bounded, i.e., the objective \eqref{encirclement}-(2) in Sec. \ref{problem} can be achieved with $\delta_3=4\bar{a}\delta_1+2t^4\check{q}$.

\begin{theorem}
If all pseudo-forces act simultaneously, the resultant pseudo-force $\Gamma_{i}$, $i\in \Phi_N$ in \eqref{eq16} is asymptotically convergent under the action of acceleration controller in \eqref{eq17}.
\end{theorem}

\textbf{Proof.}
Recalling the resultant pseudo-force $\Gamma_{i}$, $i\in \Phi_N$ in \eqref{eq16}, the Lyapunov-like candidate is chosen as
\begin{equation}\label{eq2-65}
\begin{split}
\V_{3,i}=U_{i},
\end{split}
\end{equation}
where
\begin{equation*}
\begin{split}
U_{i}=&U_{at,i}+U_{in,i}+U_{re,i},\\
U_{at,i}=&\frac{t^2}{4}(\Gamma_{at,i})^\top \Gamma_{at,i},\\
U_{in,i}=&-\sum_{\tilde{g}\in\Phi_{i,\tilde{N}}}\gamma_1\ln\{1-\frac{2\tilde{r}}{d_{i,\tilde{g}}}\}\step\{2\bar{r}_j-d_{i,\tilde{g}}\}\\
&\times\step\{d_{i,\tilde{g}}-2\tilde{r}\},\\
U_{re,i}=&-\sum_{\iota\in\Phi_{i,\bar{M},\bar{O}}}\gamma_2\ln\{1-\frac{\tilde{r}}{\hat{d}_{i,\iota}}\}\step\{\bar{r}_j-\hat{d}_{i,\iota}\}\\
&\times\step\{\hat{d}_{i,\iota}-\tilde{r}\}.
\end{split}
\end{equation*}

Here, $U_{at,i}>0$ holds since $\Gamma_{at,i}\neq 0$ and $t>0$. If $2\bar{r}_j \geq d_{i,\tilde{g}} \geq 2\tilde{r}$, then $\step\{d_{i,\tilde{g}}-2\tilde{r}\}=1$, $\step\{2\bar{r}_j-d_{i,\tilde{g}}\}=1$, and $0 \leq \frac{2\tilde{r}}{d_{i,\tilde{g}}} \leq 1$; otherwise, $\step\{2\bar{r}_j-d_{i,\tilde{g}}\}=0$ and $\step\{d_{i,\tilde{g}}-2\tilde{r}\}=0$. Thus, $U_{in,i}\geq 0$ is ensured for $\gamma_1>0$, and similarly, $U_{re,i} \geq 0$. Thus, the Lyapunov-like candidate always satisfies $\V_{3,i}>0$.

 Based on the dynamic model in \eqref{eq1-1} and \eqref{eq1-2}, the difference of $V_{3,i}$ can be obtained as

\begin{equation}\label{eq2-66}
\begin{split}
\triangle \V_{3,i}\thickapprox& \frac{\partial U_{i}}{\partial \bx_i}(\bx_i^{(+)}-\bx_i)\\
=&-(\Gamma_{at,i}+\Gamma_{in,i}+\Gamma_{re,i})^{\top}(t\bv_i+\frac{1}{2}t^2\bu_i).
\end{split}
\end{equation}

Denote two variables as $\epsilon_1=\frac{\epsilon}{\max\{\epsilon,\|\bar{\Gamma}_{in,i}\|\}}$ and $\epsilon_2=\frac{\epsilon}{\max\{\epsilon,\|\bar{\Gamma}_{re,i}\|\}}$. Considering the acceleration controller $\bu_i$ in \eqref{eq17}, the difference of $V_{3,i}$ can be further inferred as
\begin{equation}\label{eq2-67}
\begin{split}
\triangle \V_{3,i}\leq&-\frac{1}{2}t^2\Big\{(\Gamma_{at,i}+\sqrt{\epsilon_1}\Gamma_{in,i}+\sqrt{\epsilon_2}\Gamma_{re,i})^{\top}\\
&\times(\Gamma_{at,i}+\sqrt{\epsilon_1}\Gamma_{in,i}+\sqrt{\epsilon_2}\Gamma_{re,i})\\
&+(\sqrt{\epsilon_1}-1)^2\Gamma^{\top}_{at,i}\Gamma_{in,i}+(\sqrt{\epsilon_2}-1)^2\\
&\times\Gamma^{\top}_{at,i}\Gamma_{re,i}+(\sqrt{\epsilon_1}-\sqrt{\epsilon_2})^2\Gamma^{\top}_{in,i}\Gamma_{re,i}\Big\}.
\end{split}
\end{equation}

The following is to further analyze the convergence of the resultant pseudo-force $\Gamma_{i}$ from several cases.

Case 1: (Without upper limit) When none of the included angles between the pseudo-forces fall within the range of $(0^{\circ},90^{\circ})$, specifically $\angle(\Gamma_{in,i},\Gamma_{at,i})\notin (0^{\circ},90^{\circ})$, $\angle(\Gamma_{re,i},\Gamma_{at,i})\notin (0^{\circ},90^{\circ})$, and $\angle(\Gamma_{in,i},\Gamma_{re,i})\notin (0^{\circ},90^{\circ})$, the pseudo-forces satisfy $\Gamma_{in,i}=\bar{\Gamma}_{in,i}$ and $\Gamma_{re,i}=\bar{\Gamma}_{re,i}$. In this case, we can conclude that both $\epsilon_1$ and $\epsilon_2$ are equal to $1$. Therefore, the following inequality can be obtained as $\triangle \V_{3,i}
=-\frac{1}{2}t^2\Gamma^{\top}_{i}\Gamma_{i}
\leq0$.

Case 2: (With upper limit) When the angles between the pseudo-forces satisfy $
\angle(\Gamma_{in,i},\Gamma_{at,i})\cap \angle(\Gamma_{re,i},\Gamma_{in,i})\in [0^{\circ},90^{\circ}] $ or $
\angle(\Gamma_{in,i},\Gamma_{at,i})\in [0^{\circ},90^{\circ}] \cap (\Gamma_{re,i}=0)$, we can establish $\epsilon_2=1$, $\Gamma^{\top}_{at,i}\Gamma_{in,i}\geq0$ and $\Gamma^{\top}_{re,i}\Gamma_{in,i}\geq0$. Similarly, when $
\angle(\Gamma_{re,i},\Gamma_{at,i})\cap \angle(\Gamma_{re,i},\Gamma_{in,i})\in [0^{\circ},90^{\circ}]$ or $
\angle(\Gamma_{re,i},\Gamma_{at,i})\in [0^{\circ},90^{\circ}]\cap (\Gamma_{in,i}=0)$, we can known $\epsilon_1=1$, guaranteeing that both $\Gamma^{\top}_{at,i}\Gamma_{re,i}\geq0$ and $\Gamma^{\top}_{in,i}\Gamma_{re,i}\geq0$ hold true. Consequently, we can conclude that $\triangle \V_{3,i}\leq0$.

%
%

Therefore, based on the LaSalle's principle, the resultant pseudo-force $\Gamma_{i}$, $i\in \Phi_N$ is asymptotically convergent for $\V_{3,i}>0$ and $\triangle \V_{3,i}\leq0$, i.e., the objective \eqref{encirclement}-(1) in Sec. \ref{problem} can be achieved.

\begin{remark}  
In this study, the perfect encirclement of the target in the presence of obstacles cannot be guaranteed. However, based on Theorem 3, when both repulsive and interaction pseudo-forces are present, the energy field $U_{i}$ decreases, causing the drone to move away from obstacles and other drones, ensuring these forces become zero. Additionally, Theorem 2 concludes that the drones can achieve AS-based encirclement of the target.
\end{remark}

\begin{figure}
\centering
  \includegraphics[width=8cm]{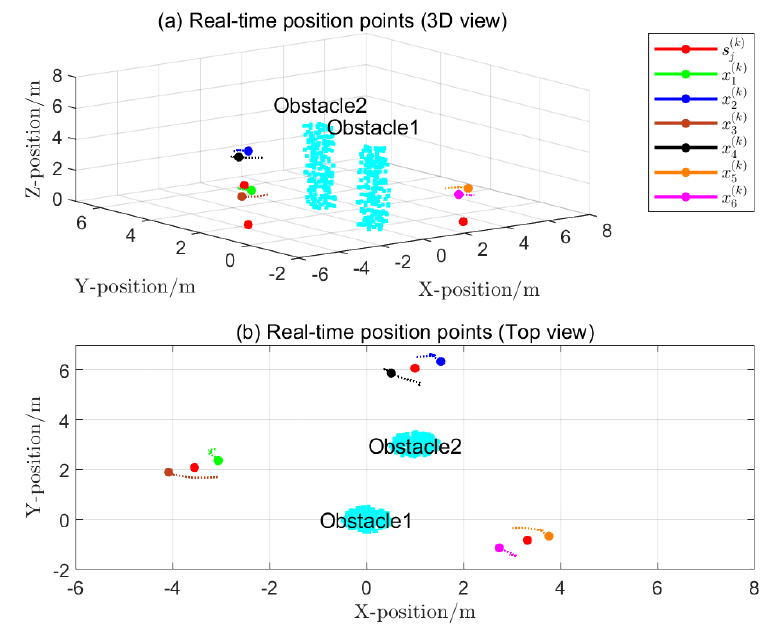}
  \vspace{-10pt}
 \caption{\footnotesize Snapshots: The real-time position points where six drones encircle three targets.}
  \label{3d}
\end{figure} 

\begin{figure}
\centering
  \includegraphics[width=8cm]{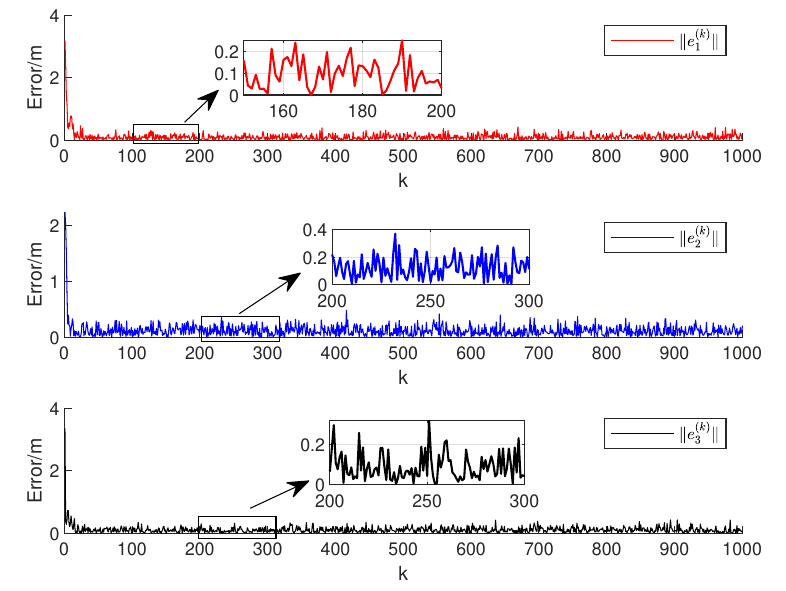}
  \vspace{-10pt}
 \caption{\footnotesize The error trajectories of DTSE.}
  \label{estimation_error}
\end{figure}

\vspace{-0.5cm}
\section{Simulation Results}
In this section, numerical simulation examples will be given to demonstrate the effectiveness of the designed algorithm. This work focuses on a specific scenario involving the UAVs pursuing and monitoring the UGVs. This scenario is a case study to assess the algorithm's performance and effectiveness in a practical application.

The initial positions of the targets and the drones in the x-axis (meter) and y-axis (meter) are given as
\begin{equation*}
\begin{split}
&\bx_1^{^{(0)}}=[1.5;2],~\bx_2^{^{(0)}}=[2;2],~\bx_3^{^{(0)}}=[2.5;2]\\
&\bx_4^{^{(0)}}=[3;2],~\bx_5^{^{(0)}}=[3.5;2],~\bx_6^{^{(0)}}=[4;2]\\
&\bs_1^{^{(0)}}=[-2;2.5],~\bs_2^{^{(0)}}=[2;1],~\bs_3^{^{(0)}}=[3;2.5].\\
\end{split}
\end{equation*}
Suppose that all drones have fixed position heights, i.e., the z-axis value of the drone is 2. The initial velocities of all targets and drones are assumed to be zero.

\begin{figure}
\centering
  \includegraphics[width=8cm]{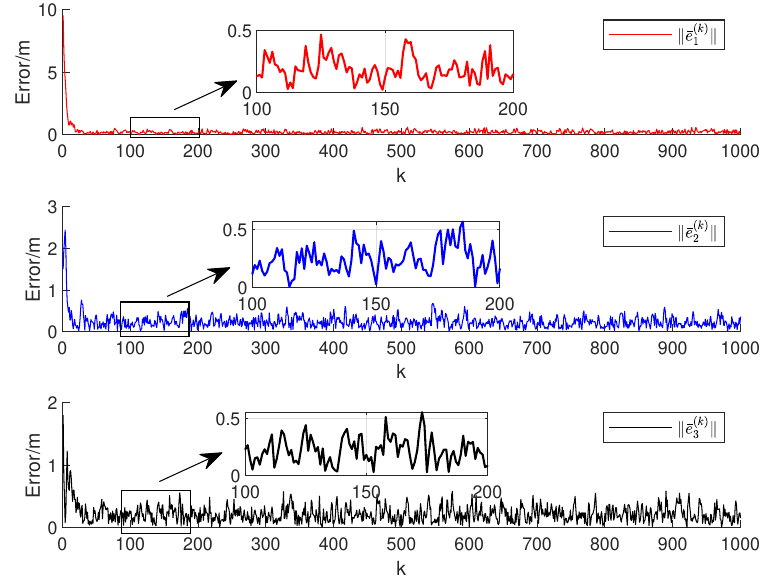}
  \vspace{-10pt}
 \caption{\footnotesize The error trajectories of AS-based encirclement}
  \label{as_error}
    \vspace{-15pt}
\end{figure}

\begin{figure}
\centering
  \includegraphics[width=8cm]{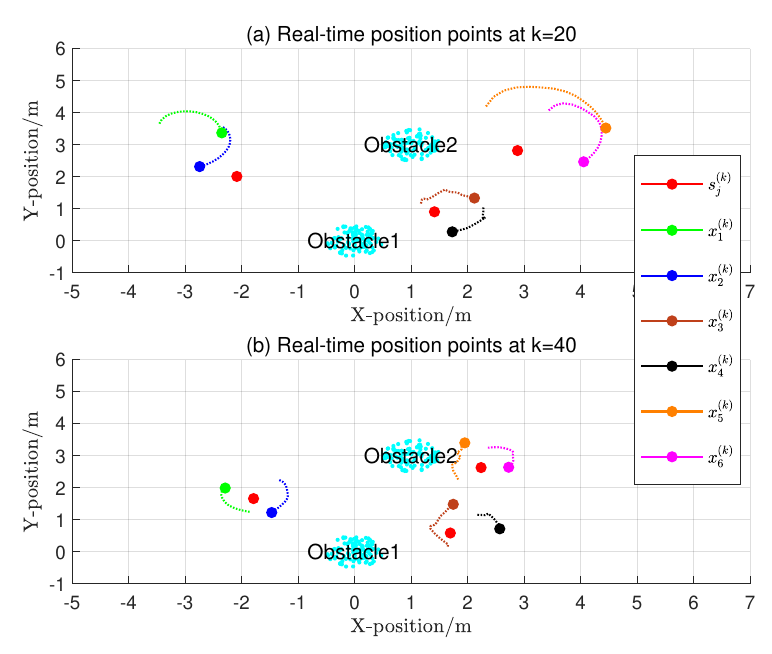}
  \vspace{-10pt}
 \caption{\footnotesize Snapshots (from top view): The real-time position points at the initial phase.}
   \vspace{-15pt}
  \label{initial_phase}
\end{figure}


Other parameters can be set as follows, $N=6$, $t=0.8$(seconds), $Q_j=[0.05,0;0,0.05] $(meter), $q_i=0.005 $(meter), $\gamma_1=0.05$ and $\gamma_2=0.005$. 
    To ensure the safety of drones and prevent collisions, the safety radius $\tilde{r}$ for the drones is set to 0.2, and $\Delta r$ in the pseudo-force action radius $\bar{r}_{j}$ is 0.1.  Furthermore, the preset trajectory shape of drone $i$ is set as $\bm{\mathcal{P}}=0.5[\sin(\frac{k\pi}{24});\cos(\frac{k\pi}{24})]$.

\begin{figure}
\centering
  \includegraphics[width=8cm]{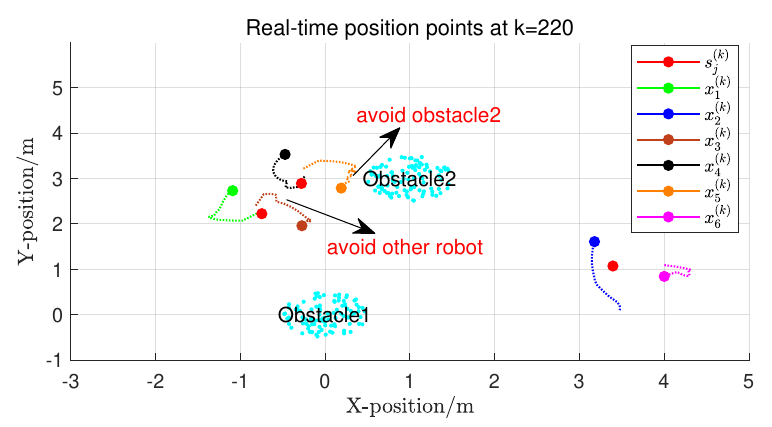}
  \vspace{-10pt}
 \caption{\footnotesize Snapshots (from top view): The real-time position points where the drones avoid obstacles.}
  \label{avoid}
\end{figure}

Using the target state estimator in \eqref{eq8-1} and the encirclement controller in \eqref{eq17}, various simulation results were generated in Matlab. Figure \ref{3d} shows both 3D and top views of the drones encircling the targets. In Figure \ref{estimation_error}, the target state estimation errors $\be_j$ converge within a small range, between 0 and 0.4 meters. Figure \ref{as_error} illustrates the rapid convergence of the AS-based encirclement error $\bar{\be}_j$, though its error threshold is larger compared to the estimation error $\be_j$. Furthermore, Real-time position snapshots at different sampling instants $k$ from a top view are shown in Figures \ref{initial_phase} to \ref{avoid}. In Figure \ref{initial_phase}(a), the initial phase illustrates that all drones are assigned targets, though encirclement is not yet achieved. Sub-figure (b) shows the drones successfully encircling the targets in an anti-synchronous manner.

\begin{figure}
\centering
  \includegraphics[width=8cm]{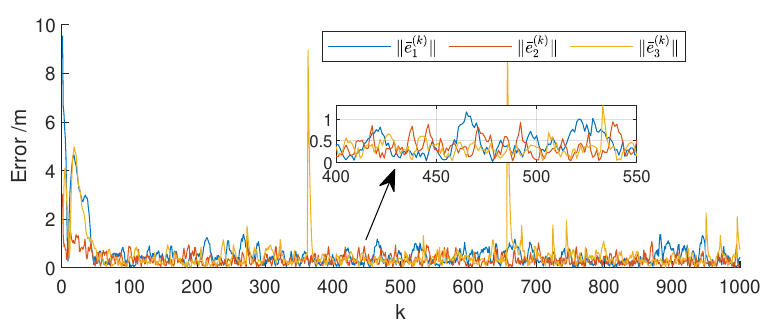}
  \vspace{-10pt}
 \caption{\footnotesize The AS-based encirclement error trajectories for the unlimited pseudo-forces.}
  \label{compare_error}
    \vspace{-15pt}
\end{figure}

\begin{figure}
\centering
  \includegraphics[width=8cm]{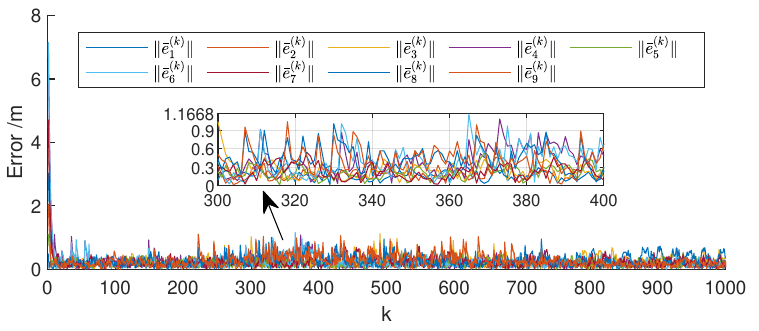}
  \vspace{-10pt}
 \caption{\footnotesize The AS-based encirclement error trajectories for nine targets.}
   \vspace{-15pt}
  \label{mutiple_error}
\end{figure}

\begin{figure}
\centering
  \includegraphics[width=8cm]{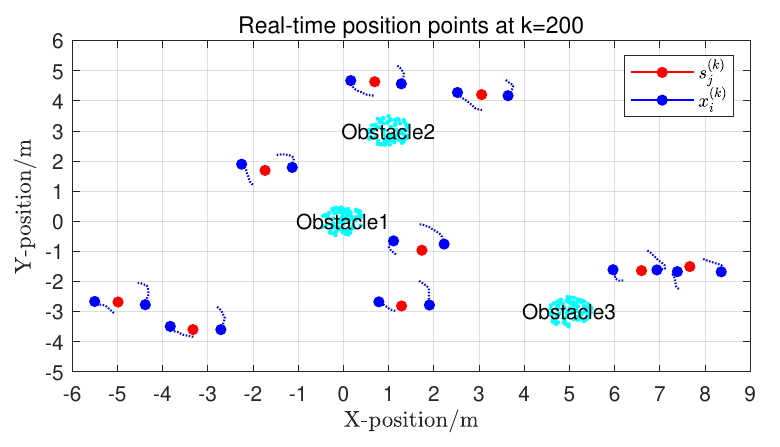}
  \vspace{-10pt}
 \caption{\footnotesize Snapshots (from top view): The real-time position points where eighteen drones encircle nine targets.}
  \label{mutiple_trajectory}
\end{figure}

Based on the pseudo-forces, the drones effectively avoid obstacles and other drones, as shown in Figure \ref{avoid}. Notably, unlike most studies, which do not impose an upper limit on pseudo-forces \cite{yan2012multilevel}, we observe in Figure \ref{compare_error} that encirclement errors show spikes at certain time instances, compared to the smoother errors in Figure \ref{as_error}. To further validate the effectiveness of the task assignment mechanism and acceleration controller, we conducted experiments with an increased number of targets. Simulation results are shown in Figures \ref{mutiple_error} and \ref{mutiple_trajectory}. Figure \ref{mutiple_trajectory} illustrates a relatively dispersed movement pattern between targets.

\section{Experimental Results}
\begin{figure}
\centering
  \includegraphics[width=7cm]{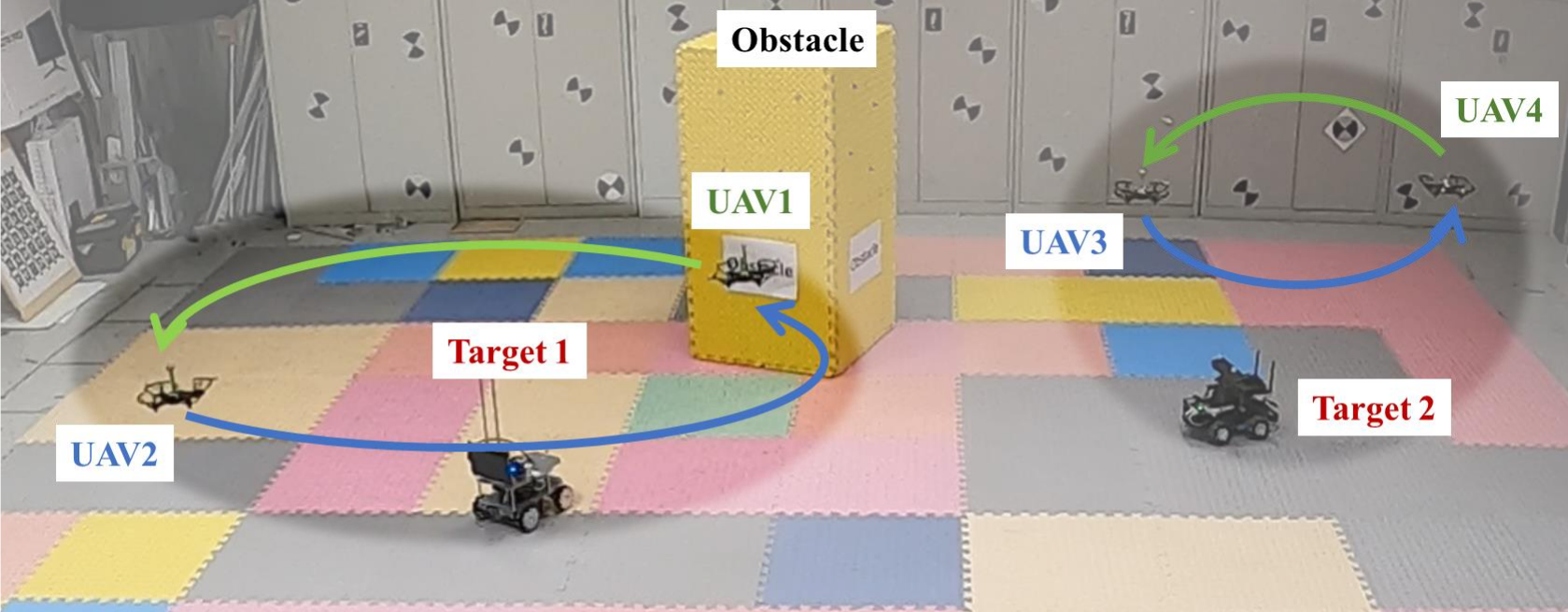}
  \vspace{-5pt}
 \caption{\footnotesize The experimental setup.}
  \label{frame}
    \vspace{-10pt}
\end{figure}

In this section, a physical experiment using the commercial off-the-shelf (COTS) low-cost Tello drones and the homologic UGVs is given to verify the validity of the designed algorithm further. As Figure \ref{frame} shows, two holonomic UGVs are treated as the targets, and four Tello drones are the tasking drones.

\begin{figure}
\centering
  \includegraphics[width=8cm]{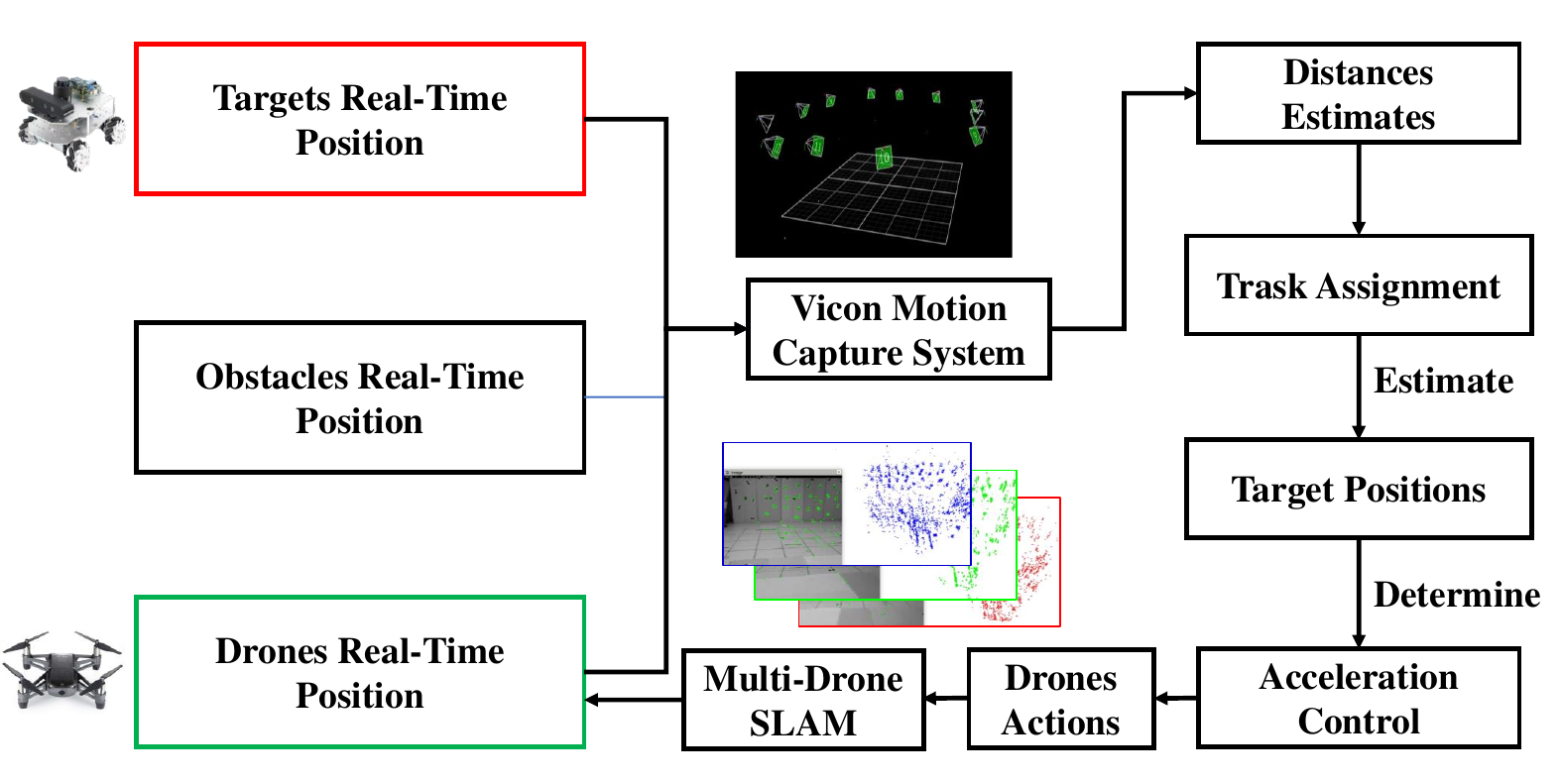}
  \vspace{-5pt}
 \caption{\footnotesize The structure of the field experiments.}
  \label{overall_experiment}
    \vspace{-15pt}
\end{figure}

\begin{figure}
\centering
  \includegraphics[width=8cm]{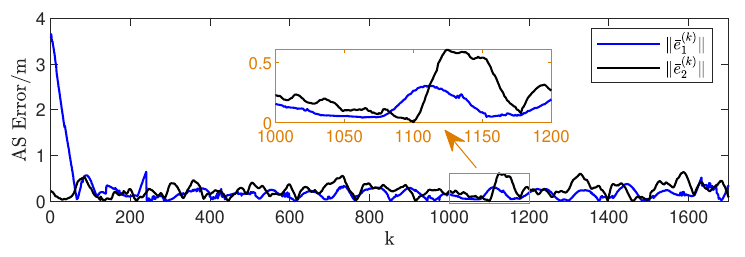}
 \caption{\footnotesize The AS error encirclement trajectories in part of experiments.}
  \label{tello_error}
\end{figure}

The entire experimental process is shown in Figure \ref{overall_experiment}. Due to the weight limitations of Tello drones, the integration of advanced sensors onboard is not feasible. Consequently, in this experiment, the motion capture system is adopted to acquire distance measurements between the drone and its target, as well as the distance between the drone and potential obstacles. To enhance the proximity of experimental outcomes to real-world outdoor applications, each drone solely relies on Vicon to obtain distance information, while all other data depends on the sensors it carries. Specifically, all drones use the equipped low-performance visual cameras and altitude sensors to construct individual SLAM systems to establish a unified coordinate system and acquire the real-time position data for all drones. The remainder of the experiment strictly follows the proposed algorithm to achieve encirclement.
Within this experiment, the distance measurement noises were not factored in, and the positions of the drones, targets, and obstacles acquired from the Vicon system were considered the ground truth. The position sampling frequency is approximately 7 Hz when four drones are running concurrently. From Figure \ref{tello_error}, the AS-based encirclement error $\|\bar{\be}_j\|$ is less than 0.5 meters, which indicates that target 1 and target 2 can be tracked and encircled effectively. Furthermore, other experiment results can be found in \url{https://youtu.be/HAtOTANfCdY}.

\vspace{-0.2cm}
\section{conclusion}
In this work, we have addressed the encirclement problem of a group of freely moving non-cooperative targets in an obstacle-rich environment. Firstly, 
through the task assignment mechanism and the real-time measured distances collected by two tasking drones, we have derived a measurement output related to the target state, wherein the impact of noise with unknown variance is present. Secondly, by incorporating the estimation of measurement output variance and the Kalman filtering technique, an innovative DTSE approach has been obtained to ascertain the positions of targets and obstacles. Thirdly, an AS-based acceleration controller has been designed to ensure that all drones can encircle all targets while avoiding obstacles. Finally, three theorems have been provided to demonstrate the stability of the entire strategy. Moreover, the effectiveness of the proposed strategy is further validated through simulation and experimental results. Additionally, a fascinating direction for future exploration involves the precise design of drone encircling trajectories to the extent that two drones can confine a single drone within a fixed spatial region.

\vspace{-0.2cm}
\section*{Appendix A}
\section*{Proof of Lemma \ref{Uniform Observability}}
Under the task assignment mechanism, drones $g^{(k)},\ldots, g^{(k-m_1)}$ are respectively collaborating with drone $i$ to encircle target $j$ in time interval $[k-m_1,k]$, and $k-4\geq m_1\geq0$.  
Recalling the expression of $A_2$ and $C_{i,g}$ in \eqref{eq3} and \eqref{eq5}, the observability Gramin matrix $\mathcal{O}_1$ of system \eqref{eq3}, defined in Chapter VI of \cite{aoki1967optimization}, can be obtained by $\mathcal{O}_1=\mathcal{O}_2^\top \hat{\Upsilon}^{-1}\mathcal{O}_2$, 
where the diagonal matrix $\hat{\Upsilon}=\diag\{\hat{\Upsilon}_{i,g^{(k)},j}^{(k)},\hat{\Upsilon}_{i,g^{(k-1)},j}^{(k-1)},\ldots,$ $\hat{\Upsilon}_{i,g^{(k-m_1)},j}^{(k-m_1)}\}$ with the expression $\hat{\Upsilon}_{i,g,j}$ in \eqref{eq7}, and the observability matrix
 \begin{equation*}
 \begin{split}
 \mathcal{O}_2=-2\left[
                   \begin{array}{cc}
                     (\bp_{i,g^{(k)}}^{(k)})^{\top}F^{\top} & 0_{1\times 2} \\
                      (\bp_{i,g^{(k-1)}}^{(k-1)})^{\top}F^{\top} & -t(\bp_{i,g^{(k-1)}}^{(k-1)})^{\top}F^{\top}  \\
                     \vdots & \vdots  \\
                     (\bp_{i,g^{(k-m_1)}}^{(k-m_1)})^{\top}F^{\top} & -m_1t(\bp_{i,g^{(k-m_1)}}^{(k-m_1)})^{\top}F^{\top}  \\
                   \end{array}
                 \right].
 \end{split}
 \end{equation*}

Recalling the model \eqref{eq1-1} and the controller \eqref{eq17}, we have $F\bp_{i,g^{(+)}}^{(+)}=F\bp_{i,g}+\frac{1}{2}t^2F(\Gamma_{i}-\Gamma_{g})$. From the resultant pseudo-forces in \eqref{eq16}, $F(\Gamma_{i}-\Gamma_{g})=-\frac{2}{t^2}(F\bp_{i, g}-2\bm{\mathcal{P}})+F\check{\Gamma}_i$, where $\check{\Gamma}_i=\bar{\Gamma}_{in,i}+\bar{\Gamma}_{re,i}-\bar{\Gamma}_{in,g}-\bar{\Gamma}_{re,g}$. 
 According to Assumption \ref{without_repulsive_force}, there always exist four consecutive instants $\{k_1+1,k_1,k_1-1,k_1-2\}$ within the time interval $[k-m_1, k]$ such that $\check{\Gamma}_i$ equals zero. Based on this, we have $F(\Gamma_{i}-\Gamma_{g})=-\frac{2}{t^2}(F\bp_{i, g}-2\bm{\mathcal{P}})$. Furthermore, we can obtain $F\bp_{i,g^{(+)}}^{(+)}=2\bm{\mathcal{P}}$. Below, we will assess whether $\mathcal{O}_2$ attains full column rank, offering insights into the observability of the system.

Based on the definition of $\bm{\mathcal{P}}$ in \eqref{eq4} and Assumption 6, we transform any four consecutive rows $\{k_1+1, k_1, k_1-1, k_1-2\}$ of $\mathcal{O}_2$ as $\tilde{\mathcal{O}}_2$.
 \begin{equation*}
 \begin{split}
 \tilde{\mathcal{O}}_2=-4\rho\left[
                   \begin{array}{cccc}
                     \tilde{o}_{11}&\tilde{o}_{12}&k_2t\tilde{o}_{11}&k_2t\tilde{o}_{12}\\
                     \tilde{o}_{21}&\tilde{o}_{22}&(k_2-1)t\tilde{o}_{21}&(k_2-1)t\tilde{o}_{22}\\
                    \tilde{o}_{31}&\tilde{o}_{32}&(k_2-2)t\tilde{o}_{31}&(k_2-2)t\tilde{o}_{32}\\
                    \tilde{o}_{41}&\tilde{o}_{42}&(k_2-3)t\tilde{o}_{41}&(k_2-3)t\tilde{o}_{42}\\
                   \end{array}
                 \right],
 \end{split}
 \end{equation*}
 where $k_2=k_1+1-k$. Moreover,
  \begin{equation*}
 \begin{split}
\tilde{o}_{11}&=\sin(\bar{\nu}k_1\pi),~\tilde{o}_{12}=\cos(\bar{\nu}k_1\pi),\\
\tilde{o}_{21}&=\sin(\bar{\nu}(k_1-1)\pi),~\tilde{o}_{22}=\cos(\bar{\nu}(k_1-1)\pi),\\
\tilde{o}_{31}&=\sin(\bar{\nu}(k_1-2)\pi),~
\tilde{o}_{32}=\cos(\bar{\nu}(k_1-2)\pi),\\
\tilde{o}_{41}&=\sin(\bar{\nu}(k_1-3)\pi),
~\tilde{o}_{42}=\cos(\bar{\nu}(k_1-3)\pi).
 \end{split}
 \end{equation*}

By contradiction, assume there exist some real constants $\mu_1$, $\mu_2$, $\mu_3$ and $\mu_4$ such that the following hold.
\begin{subequations}\label{eq38}
\begin{align}
&\mu_1\tilde{o}_{11}+\mu_3\tilde{o}_{31}+2\mu_4\tilde{o}_{41}=0,\label{eq38-1}\\
&\mu_1\tilde{o}_{12}+\mu_3\tilde{o}_{32}+2\mu_4\tilde{o}_{42}=0,\label{eq38-2}\\
&\mu_2\tilde{o}_{21} +2\mu_3\tilde{o}_{31}+3\mu_4\tilde{o}_{41}=0,\label{eq38-3}\\
&\mu_2\tilde{o}_{22}+2\mu_3\tilde{o}_{32}+3\mu_4\tilde{o}_{42}=0. \label{eq38-4}
 \end{align}
 \end{subequations}

Based on the first two equations \eqref{eq38-1} and \eqref{eq38-2}, we have equation \eqref{eq40-1}. Then, based on the later two equations \eqref{eq38-3} and \eqref{eq38-4}, we have equation \eqref{eq40-2}.
\begin{subequations}\label{eq40}
\begin{align}
& \mu_3=-\frac{2\sin(3\bar{\nu}\pi)}{\sin(2\bar{\nu}\pi)}\mu_4,\label{eq40-1}\\
& \mu_4=-\frac{1}{3\cos(\bar{\nu}\pi)}\mu_3. \label{eq40-2}
 \end{align}
 \end{subequations}

Considering the frequency of circumnavigation $\bar{\nu}=\frac{1}{\ell}$ with $\ell\geq4$, it is easy to obtain $\mu_3=0$ and $\mu_4=0$ for $\cos(\bar{\nu}\pi)\neq0$, $\sin(2\bar{\nu}\pi)\neq0$ and $\sin(3\bar{\nu}\pi)\neq0$. Recalling $\eqref{eq38-1}$ and $\eqref{eq38-3}$, it follows that $\mu_1$, $\mu_2$, $\mu_3$, and $\mu_4$ are all equal to zero, indicating that the rows of the matrix $\tilde{\mathcal{O}}_2$ are linearly independent. Therefore, the matrix $\tilde{\mathcal{O}}_2$ is full rank.

Consequently, based on the above analysis, we can consider that the observation matrix $\mathcal{O}_2$ is full column rank within the time interval $[k-m_1,k]$. Therefore, $\mathcal{O}_2^\top \mathcal{O}_2$ is positive matrix.
Based on the expression of $\mathcal{O}_2$, $\lambda_{\min}\{\mathcal{O}_2^\top \mathcal{O}_2\}>0$ and $\lambda_{\max}\{\mathcal{O}_2^\top \mathcal{O}_2\}$ always can be obtained.

Furthermore, in terms of [Theorem 3.1 , \cite{feng2014kalman}], the estimated variance $\hat{\Upsilon}_{i,\imath,j}$ is approximately equal to the actual variance $\Upsilon_{i,\imath,j}$. Referring to the formula \eqref{eq18-3} from Assumption \ref{limit}, we can estimate that $\hat{r}I_{(m+1)\times(m+1)}\leq\hat{\Upsilon}\leq\check{r}I_{(m+1)\times(m+1)}$.

Therefore, we have $\Lambda_1I_{4\times4}\leq\mathcal{O}_1\leq\Lambda_2I_{4\times4}$ with $\Lambda_1=\hat{r}\lambda_{\min}\{\mathcal{O}_2^\top \mathcal{O}_2\}>0$ and $\Lambda_2=\check{r}\lambda_{\max}\{\mathcal{O}_2^\top \mathcal{O}_2\}$, which imply that the state model of target $j$ in \eqref{eq3} is uniformly observable \cite {anderson1981detectability}.


\section*{Appendix B}
\section*{Proof of Lemma \ref{variance bound}}
Utilizing the matrix inversion lemma and the
target position estimator gain $K_{j}$ in \eqref{eq8-5}, the estimation error variance \eqref{eq8-3} can be rewritten as
\begin{equation}\label{eq2-42}
\begin{split}
\bm{\zeta}^{-1}_j=(\bm{\zeta}^{(k|k-1)}_j)^{-1}+C_{i,g}^{\top}\hat{\Upsilon}_{i,g,j}^{-1}C_{i,g}.
\end{split}
\end{equation}

By recalling the formula \eqref{eq8-4}, the term $B_2Q_{j} B_2^\top$ is irreversible for the minimum eigenvalue $\underline{b}=0$. Therefore,
we can only rephrase the equation \eqref{eq2-42} as:
\begin{equation}\label{eq2-43}
\begin{split}
\bm{\zeta}^{-1}_j\leq (A_2\bm{\zeta}_j^{(-)}A_2^\top )^{-1}+C_{i,g}^{\top}\hat{\Upsilon}_{i,g,j}^{-1}C_{i,g}.
\end{split}
\end{equation}

For $\lambda_{\max}\{C_{ij}^{\top}C_{ij}\}=4\lambda_{\max}\{\bp_{ij}\bp_{ij}^{\top}\}$, we have $ C_{ij}^{\top}C_{ij}\leq 4\check{c}I_{4 \times 4}$. Therefore, considering Assumption \ref{limit} and $\lambda_{\min}\{A_2A_2^\top\}=\underline{a}$, we have $\bm{\zeta}_j^{-1}\leq \big\{\underline{a}^{-k}(\lambda_{\min}\{\bm{\zeta}_j^{(0)}\})^{-1}+\sum_{\imath=0}^{k-1}4\underline{a}^{-\imath}\check{c}\hat{r}^{-1}\big\}I_{4 \times 4}$. Since $0 < \underline{a} < 1$, $\underline{a}^{-k}\to \infty$ as $k \to \infty $. Therefore, we analyze the upper bounds of $\bm{\zeta}^{-1}_j$ in two cases.

Case 1: When $0<k<m_1$, building upon \eqref{eq2-43}, we deduce the inequality $\bm{\zeta}_j^{-1}\leq \check{\zeta}_j'I_{4\times 4}$ with $\check{\zeta}_j'=\underline{a}^{1-m_1}(\lambda_{\min}\{\bm{\zeta}_j^{(0)}\})^{-1}+\sum_{\imath=0}^{m_1-2}4\underline{a}^{-\imath}\check{c}\hat{r}^{-1}$.

Case 2: When $k\geq m_1$, let $W_j=(\bm{\zeta}^{(k|k-1)}_j)^{-1}$ and $W_j^{(k|k-1)}=A_2^{-T}\bm{\zeta}_j^{-1}A_2^{-1}$. Based on the formulas \eqref{eq8-4} and \eqref{eq2-42}, we have
{\begin{subequations}\label{eq2-46}
\begin{align}
W_j=&
\big\{(W_j^{(k-1|k-2)})^{-1}+B_2Q_{j} B_2^\top \big\}^{-1},\label{eq2-46-1}\\
W_j^{(k/k-1)}=&A_2^{-T}W_jA_2^{-1}+A_2^{-T}C_{i,g}^{T}\hat{\Upsilon}_{i,g,j}^{-1}C_{i,g}A_2^{-1}.\label{eq2-46-2}
 \end{align}
 \end{subequations}

 In the same manner as in [\cite{deyst1968conditions} , section III] , we further have 
\begin{equation} \label{eq2-47}
\begin{split}
W_j\leq&\big\{\sum^{k}_{\varrho=k-m_1-1}\mathcal{A}^{-\varrho}B_2Q_jB_2^\top (\mathcal{A}^{-\varrho})^{T}\big\}^{-1}\\
&+\sum^{k-1}_{\varrho=k-m_1}(\mathcal{A}^{\varrho})^{T}(C_{i,g}^{(\varrho)})^{T}(\hat{\Upsilon}_{i,g,j}^{(\varrho)})^{-1}C_{i,g}^{(\varrho)}\mathcal{A}^{\varrho},
\end{split}
\end{equation}
where $\mathcal{A}^{k}=I_{4\times 4}$, $\mathcal{A}^{\varrho}=A_2^{-(k-\varrho)}$.

Based on Lemma \ref{Uniform Observability} and Lemma \ref{Controllable}, we have $\sum^{k}_{\varrho=k-m_1-1}\mathcal{A}^{-\varrho}B_2Q_jB_2^\top (\mathcal{A}^{-\varrho})^{T}=\mathcal{H}_2\hat{Q}_j\mathcal{H}_2^\top $ and $\sum^{k}_{\varrho=k-m_1}(\mathcal{A}^{\varrho})^{T}$ $(C_{i,g}^{(\varrho)})^{T}(\hat{\Upsilon}_{i,g,j}^{(\varrho)})^{-1}C_{i,g}^{(\varrho)}\mathcal{A}^{\varrho}=\mathcal{O}_2^\top \hat{\Upsilon}^{-1}\mathcal{O}_2$. Using \eqref{eq2-42}, the following formula can be obtained,
\begin{equation*}
\begin{split}
\bm{\zeta}^{-1}_j\leq&\{\mathcal{H}_2\hat{Q}_j\mathcal{H}_2^\top \}^{-1}+\mathcal{O}_2^\top \hat{\Upsilon}^{-1}\mathcal{O}_2\leq(\frac{1}{\Lambda_3}+\Lambda_2)I_{4\times4}.
\end{split}
\end{equation*}

Therefore, the upper bound of $\bm{\zeta}_j^{-1}$ can be obtain as $\check{\zeta}_j=\max\big\{(\frac{1}{\Lambda_3}+\Lambda_2)I_{4\times4},\check{\zeta}_j'\big\}>0$.

Next, utilizing the formulas \eqref{eq2-42} and \eqref{eq8-4}, we can derive the following inequality:
\begin{equation*}
\begin{split}
\bm{\zeta}^{-1}_j\geq&(A_2\bm{\zeta}_j^{-}A_2^\top +B_2Q_{j} B_2^\top )^{-1}.
\end{split}
\end{equation*}

Considering Assumption \ref{limit} again and $\lambda_{\max}\{A_2A_2^\top\}=\bar{a}$ and $\lambda_{\max}\{B_2B_2^\top\}=\bar{b}$,  we have $
\bm{\zeta}^{-1}_j\geq\big\{\bar{a}^k\lambda_{\max}\{\bm{\zeta}_j^{(0)}\}+(\sum_{\imath=0}^{k-1}\bar{a}^\imath)\bar{b}\check{q}\big\}^{-1}I_{4\times4}$. Here, $\bar{a}>1$ always holds for $\bar{a}=\frac{2+t^2+ t\sqrt{4+t^2}}{2}$ and $t>0$. As $k\to \infty$, $\bar{a}^{k}\to \infty$. Then, we discuss the lower bound of $\bm{\zeta}^{-1}_j$ under two cases. 

Case 1: When $0<k<m_1$, we can establish the lower bound for $\bm{\zeta}^{-1}_j$ as:
\begin{equation*}
\begin{split}
\bm{\zeta}^{-1}_j\geq\hat{\bm{\zeta}}_j'I_{4\times4},
\end{split}
\end{equation*}
where $\hat{\bm{\zeta}}_j'=(\bar{a}^{m_1-1}\lambda_{\max}\{\bm{\zeta}_j^{(0)}\}+\sum_{\imath=0}^{m_1-2}\bar{a}^\imath\bar{b}\check{q})^{-1}>0$.

Case 2: When $k\geq m_1$, considering the formulas \eqref{eq2-42} and \eqref{eq8-4} again, along with the insights from Lemma \ref{Uniform Observability} and Lemma \ref{Controllable}, and based on the similar derivation of \eqref{eq2-47}, we have 
\begin{equation*}
\begin{split}
\bm{\zeta}_j&\leq \{\mathcal{O}_2^\top \hat{\Upsilon}^{-1}\mathcal{O}_2\}^{-1}+\mathcal{H}_2\hat{Q}_j\mathcal{H}_2^\top \leq (\frac{1}{\Lambda_1}+\Lambda_4)I_{4\times4}.
\end{split}
\end{equation*}

Therefore, the lower bound of $\bm{\zeta}_j^{-1}$ can be established as $\hat{\zeta}_j=\min\big\{ (\frac{\Lambda_4}{1+\Lambda_4\Lambda_1})I_{4\times4},\hat{\bm{\zeta}}_j'\big\}>0$.

\makeatletter
\let\myorg@bibitem\bibitem
\def\bibitem#1#2\par{%
	\@ifundefined{bibitem@#1}{%
		\myorg@bibitem{#1}#2\par
	}{%
		\begingroup
		\color{\csname bibitem@#1\endcsname}%
		\myorg@bibitem{#1}#2\par
		\endgroup
	}%
}
\bibliographystyle{ieeetr}
\bibliography{reference}
\begin{IEEEbiography}
[{\includegraphics[width=1in,height=1.25in]{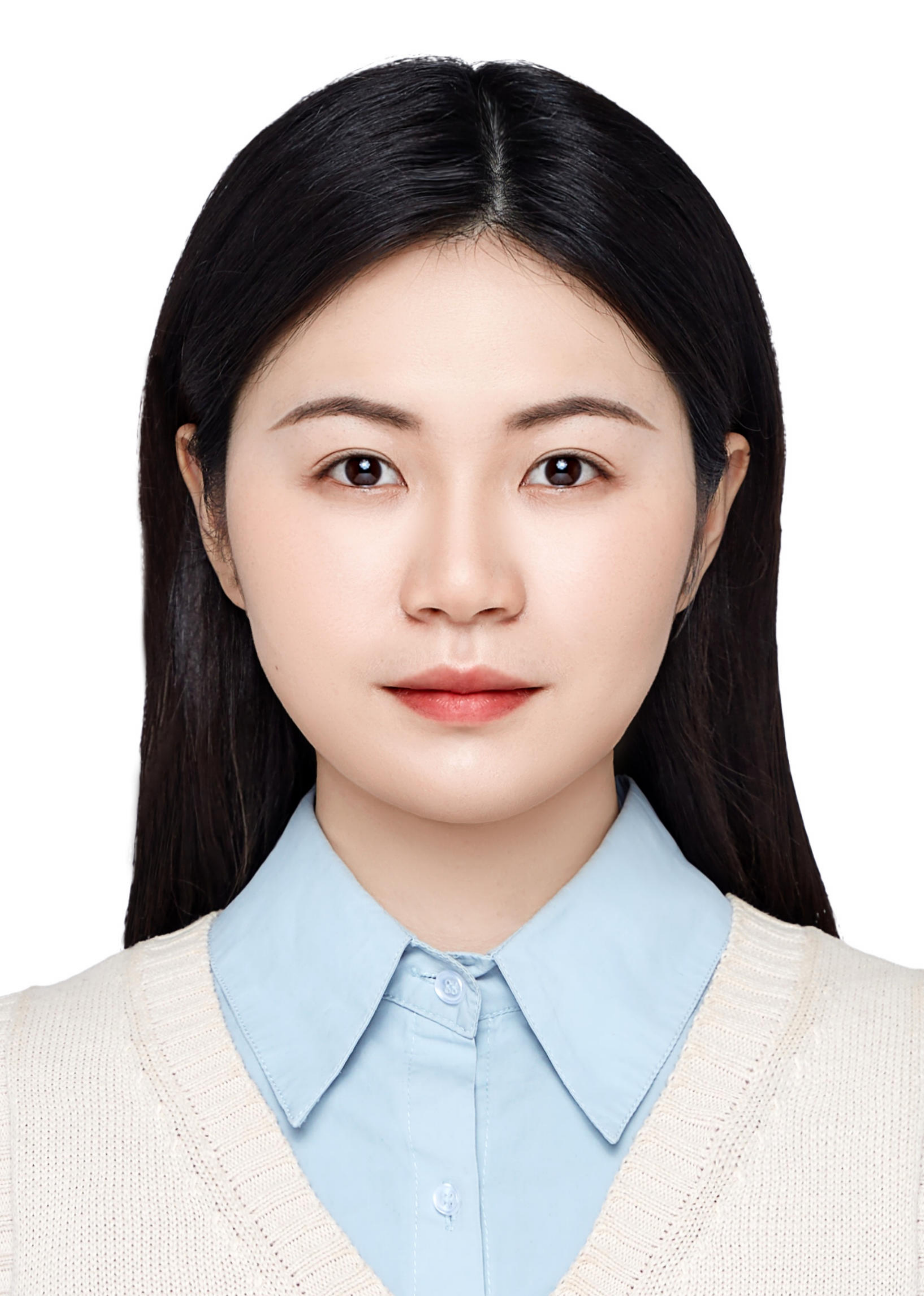}}]{Fen Liu} received the M.S. and Ph.D. degrees in 2020 and 2023, respectively, from the School of Automation, Guangdong University of Technology, Guangzhou, China. She is currently a research fellow at the School of Electrical and Electronic Engineering, Nanyang Technological University, Singapore. Her research interests include robust estimation, target encirclement, cooperative control, and anti-synchronization control.
\end{IEEEbiography}
\begin{IEEEbiography}
[{\includegraphics[width=1in,height=1.25in]{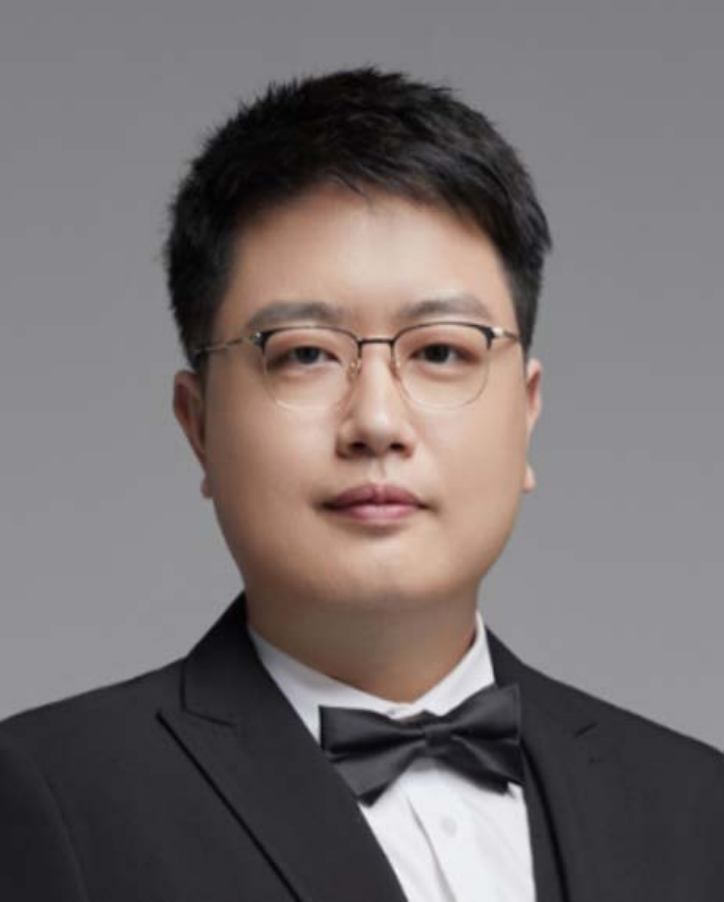}}]{Shenghai Yuan} received his B.S. and Ph.D. degrees in Electrical and Electronic Engineering in 2013 and 2019, respectively, from Nanyang Technological University, Singapore. His research focuses on robotics perception and navigation. He is a postdoctoral senior research fellow at the Centre for Advanced Robotics Technology Innovation (CARTIN), Nanyang Technological University, Singapore. He has contributed over 80 papers to journals such as TRO, IJRR, ISPRS, TIE, RAL, and conferences including ICRA, CVPR, ICCV, NeurIPS, and IROS. Currently, he serves as an associate editor for the Unmanned Systems Journal and as a guest editor of the Electronics Special Issue on Advanced Technologies of Navigation for Intelligent Vehicles. He achieved second place in the academic track of the 2021 Hilti SLAM Challenge, third place in the visual-inertial track of the 2023 ICCV SLAM Challenge, and won the IROS 2023 Best Entertainment and Amusement Paper Award. He also received the Outstanding Reviewer Award at ICRA 2024. He organized the CARIC UAV Swarm Challenge and Workshop at CDC 2023, the UG2 Anti-drone Challenge at CVPR 2024, and the second CARIC UAV Swarm Challenge at IROS 2024.  \\
\end{IEEEbiography}

\begin{IEEEbiography}
[{\includegraphics[width=1in,height=1.25in]{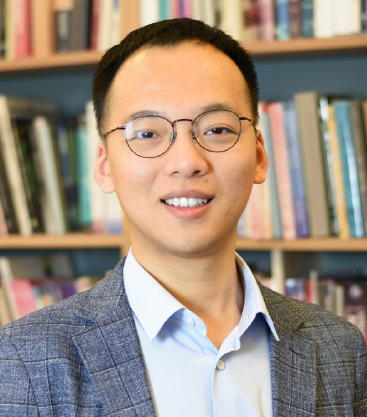}}]{Kun Cao} received the B.Eng. degree in mechanical engineering from Tianjin University, Tianjin, China, in 2016, and the Ph.D. degree in electrical and electronic engineering from the School of Electrical and Electronic Engineering, Nanyang Technological University, Singapore, in 2021. He was the 2022 Wallenberg-NTU Presidential Postdoctoral Fellow with the School of Electrical and Electronic Engineering at Nanyang Technological University and the School of Electrical Engineering and Computer Science at KTH Royal Institute of Technology in Sweden. In 2024, he is currently a faculty member in Tongji University, Shanghai, China. His research interests include individual and swarm intelligence. \\
\end{IEEEbiography}

\begin{IEEEbiography}
[{\includegraphics[width=1in,height=1.25in]{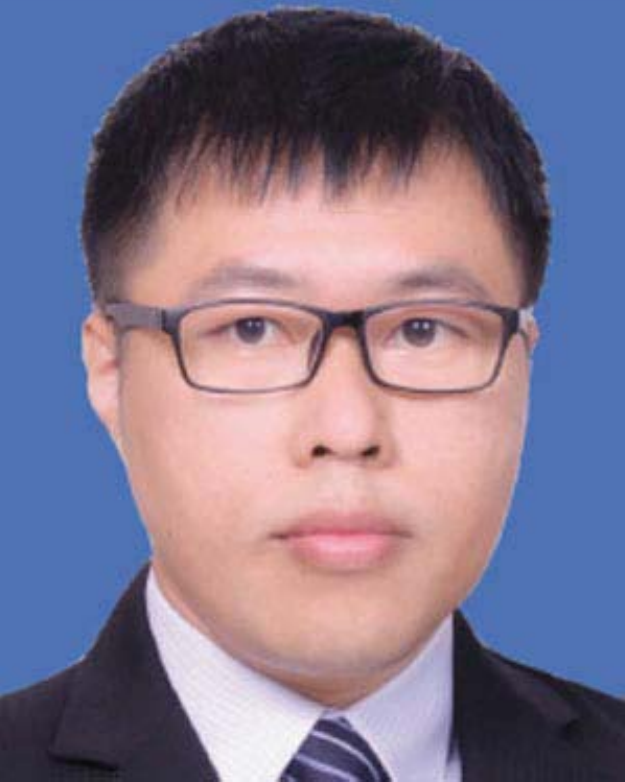}}]{Wei Meng} received the B.E. and M.E. degrees from Northeastern University, Shenyang, China, in 2006 and 2008, respectively, and the Ph.D. degree in electrical and electronic engineering from the Nanyang Technological University, Singapore, in 2013. From 2012 to 2017, he was a Research Scientist with UAV Research Group, Temasek Laboratories, National University of Singapore. He is currently a professor at Guangdong University of Technology. His research interests include unmanned systems, cooperative control, multi-robot systems, localization, and tracking.
\end{IEEEbiography}

\begin{IEEEbiography}
[{\includegraphics[width=1in,height=1.25in]{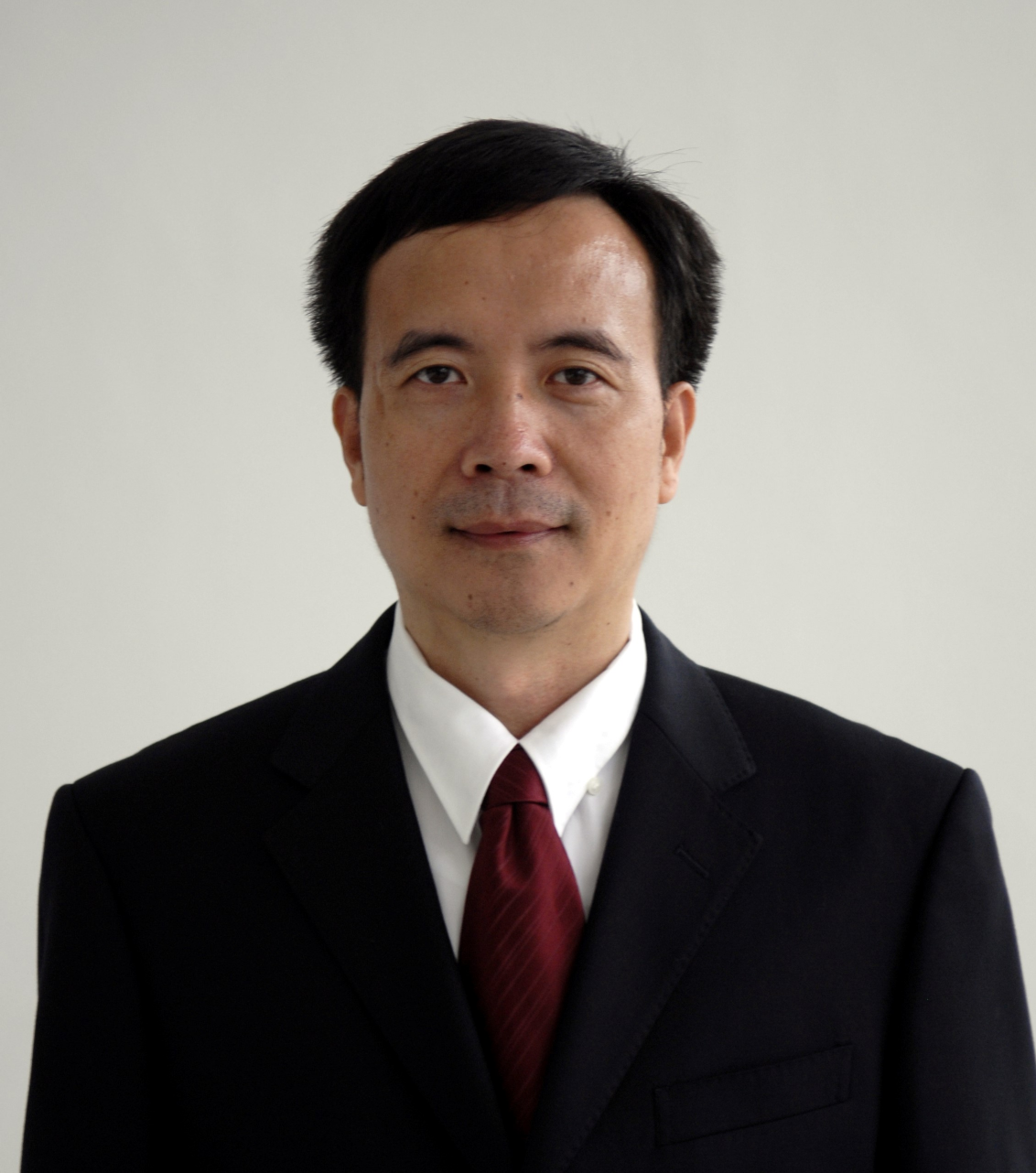}}]{Lihua Xie} (Fellow, IEEE) received the Ph.D. degree in electrical engineering from the University of Newcastle, Australia, in 1992. Since 1992, he has been with the School of Electrical and Electronic Engineering, Nanyang Technological University, Singapore, where he is currently President’s Chair in control engineering and Director, Center for Advanced Robotics Technology Innovation. He has served as the Head of Division of Control and Instrumentation and Co-Director, Delta-NTU Corporate Lab for Cyber-Physical Systems. He held teaching appointments in the Department of Automatic Control, Nanjing University of Science and Technology from 1986 to 1989. \\
Dr Xie’s research interests include robust control and estimation, networked control systems, multi-agent networks, and unmanned systems. He has published 10 books and numerous papers in the areas, and holds 25 patents/technical disclosures. He was listed as a highly cited researcher by Thomson Routers and Clarivate Analytics. He is an Editor-in-Chief for Unmanned Systems and has served as Editor for IET Book Series in Control and Associate Editor for a number of journals including IEEE Transactions on Automatic Control, Automatica, IEEE Transactions on Control Systems Technology, IEEE Transactions on Network Control Systems, etc. He was an IEEE Distinguished Lecturer (Jan 2012 – Dec 2014) and the General Chair of the 62nd IEEE Conference on Decision and Control (CDC 2023). He is currently Vice-President of IEEE Control System Society. Dr Xie is Fellow of Academy of Engineering Singapore, IEEE, IFAC, and CAA. 
\end{IEEEbiography}
\end{document}